\documentclass{article}

\usepackage[final]{neurips_2022}

\usepackage[utf8]{inputenc} 
\usepackage[T1]{fontenc}    
\usepackage{hyperref}       
\usepackage{url}            
\usepackage{booktabs}       
\usepackage{amsfonts}       
\usepackage{nicefrac}       
\usepackage{microtype}      
\usepackage{subfigure}
\usepackage[table]{xcolor}        
\usepackage{multirow}
\usepackage{amsmath}
\usepackage{todonotes}
\usepackage{ulem}
\usepackage[english]{babel}
\usepackage{amsthm}
\usepackage{adjustbox}
\usepackage[symbol*]{footmisc}

\newtheorem{theorem}{Theorem}
\newtheorem{proposition}[theorem]{Proposition}
\newtheorem*{remark}{Remark}

\newcommand{\Pool}{\mathcal{D}_\text{pool}}
\newcommand{\Val}{\mathcal{D}_\text{val}}
\title{Active Learning with Expected Error Reduction}

\setfnsymbol{bringhurst}

\newcommand{\steve}[1]{{\color{red}{Steve: #1}}}

\author{%
  Stephen Mussmann\thanks{Work done while at Apple} \\
  University of Washington \\
  \texttt{mussmann@cs.washington.edu}\\
  \And
  Julia Reisler\footnotemark[1] \\
  Stanford University \\
  \texttt{jreisler@stanford.edu}\\
  \And
  Daniel Tsai \\
  Apple\\
  \texttt{daniel\_tsai@apple.com}\\
  \And
  Ehsan Mousavi \\
  Apple\\
  \texttt{emousavi@apple.com}\\
  \And
  Shayne O’Brien \\
  Apple\\
  \texttt{shayne\_obrien@apple.com}\\
  \And
  Moises Goldszmidt\\
  Apple\\
  \texttt{mgoldszmidt@apple.com}\\
}

\begin{document}

\maketitle

\begin{abstract}
  Active learning has been studied extensively as a method for efficient data collection. Among the many approaches in literature, Expected Error Reduction (EER)~\cite{roy2001toward} has been shown to be an effective method for active learning: select the candidate sample that, in expectation, maximally decreases the error on an unlabeled set. However, EER requires the model to be retrained for every candidate sample and thus has not been widely used for modern deep neural networks due to this large computational cost. In this paper we reformulate EER under the lens of Bayesian active learning and derive a computationally efficient version that can use any Bayesian parameter sampling method (such as~\cite{gal2016dropout}). We then compare the empirical performance of our method using Monte Carlo dropout for parameter sampling against state of the art methods in the deep active learning literature. Experiments are performed on four standard benchmark datasets and three WILDS datasets~\citep{koh2021wilds}. The results indicate that our method outperforms all other methods except one in the data shift scenario -- a model dependent, non-information theoretic method that requires an order of magnitude higher computational cost~\citep{ash2019deep}.

\comment{Active learning has been studied extensively as a method for data collection. Among the many approaches in the literature, Expected Error Reduction (EER)~\cite{roy2001toward} has been shown to be effective in the selection of samples to label: select the sample that, in expectation, maximally decreases the error on an unlabeled set. However, EER requires the model be retrained for every candidate, and thus has not been widely used for modern deep neural networks due to this large computational cost. In this paper, we take a new look at EER under the lens of Bayesian active learning and derive a computationally efficient version that can use any Bayesian parameter sampling method. Furthermore, we formally compare our selection criteria to that used by other methods based on information theoretic considerations such as BALD. We then empirically compare the performance of our method, using Monte Carlo dropout for parameter sampling, against state of the art methods in the deep active learning literature, on four standard benchmarks and three WILDS datasets, each with and without data shift, for a total of 14 data settings. Our results indicate that our method outperforms all other methods except one, and only in the data shift scenario: a model dependent, non-information theoretic method that requires an order of magnitude higher computational cost.}
\end{abstract}


\section{Introduction}

Active learning studies how adaptivity in data collection can decrease the data requirement of machine learning systems \citep{settles2009active} and has been applied to a variety of applications such as medical image analysis where unlabeled samples are plentiful but labeling via annotation or experimentation is expensive. In such cases, intelligently selecting samples to label can dramatically reduce the cost of creating a training set. A range of active learning algorithms have been proposed in recent decades \citep{lewis1994sequential,tong2001support,roy2001toward}. Current research trends have shifted towards active learning for deep learning models \citep{sener2018active,ash2019deep,gal2017deep, sinha2019variational}, which present a new set of constraints that traditional methods do not always address.

Another application of active learning has been in addressing data shift \citep{rabanserFL2019, schrouffMaintaining2022}, a problem which can be summarized as deploying a model to a distribution that is different from the training distribution, and is characteristic of real world machine learning applications. This is a critical aspect of model deployment because distributional shifts can cause significant degradation in model performance. Active learning has been proposed as a method to mitigate the effects of data shift by identifying salient out-of-distribution samples to add to the training data (see, for example, \cite{kirsch2021Test} and \cite{zhaoLAY21}). 

In this paper we propose a new derivation of the Expected Error Reduction (EER) active learning method~\citep{roy2001toward} and apply it to deep neural networks in experiments with and without data shift, expanding upon the original setting of simple classifiers such as Naive Bayes. At its core, EER chooses a sample that maximally reduces the generalization "error" or "loss" in expectation. The original implementation of EER requires retraining the classifier model for every possible candidate sample, for every possible label. The cost of this constant retraining is intractable in the deep neural network context. To circumvent this, we present a formulation that estimates the value of a label for reducing the classifier's loss without retraining the neural network. We achieve this by casting EER in the context of Bayesian methods and drawing samples from the model posterior conditioned on the data.  Although sampling from the Bayesian posterior of model parameters directly can be difficult, there are a variety of methods to approximate this distribution such as Monte Carlo dropout \citep{gal2016dropout}, ensembles \citep{beluch2018power}, and Langevin dynamics \citep{welling2011bayesian}. 

Perhaps the earliest Bayesian data collection method is Expected Information Gain \citep{lindley1956measure} which is equivalent to selecting the sample with the label that has the highest mutual information with the model parameters.  This method has been explored by \citet{gal2017deep} for active learning in the context of deep neural networks, and is referred to as BALD. While BALD maximizes the information encoded by model parameters about the dataset, the fundamental goal of classification is to identify the model parameters that minimize test error. This imperfect alignment in the objective is what motivates our exploration of EER. Similar to \citet{roy2001toward}, we consider two variants of the method: one based on zero-one loss, which we refer to as minimization of zero-one loss (MEZL), and the other on log-likelihood loss, which we refer to as minimization of log-likelihood loss (MELL). Our derivation also enables us to cast EER in an information theoretic setting that allows for a theoretical comparison with BALD. Both BALD and MELL reduce the epistemic uncertainty, but MELL additionally removes task irrelevant information (see~\ref{sec:decomposition}). We hypothesize that this explains why MELL outperforms BALD empirically (see Section~\ref{sec:results}).

In our experiments we compare MELL and MEZL against state of the art active learning methods including BADGE~\citep{ash2019deep}, BALD~\citep{gal2017deep}, and Coreset~\citep{sener2018active} on seven image classification datasets. For each dataset, we perform experiments in both a setting with and without data shift.  We remark that unlike the  work of \cite{kirsch2021Test} where the goal is to avoid labeling out of distribution samples, we are interested in the realistic scenario of reducing model error on a test set that may be drawn from a different distribution than the seed set. In our results, MELL consistently performs better or on par with all other approaches in both the no data shift and data shift settings, with the largest gains above BALD coming in the data shift setting. 

The only approach that compares favorably to ours is BADGE, which outperforms MELL and MEZL on two datasets under data shift: Camelyon17 and MNIST. However, MELL and MEZL require orders of magnitude less computation cost (see Section ~\ref{sec:results} and Appendix~\ref{sec:computationalComplexity}) and are model agnostic -- they can be applied to neural networks, random forests, linear models, etc.
We also remark that optimizing the log likelihood (MELL) performs better than zero-one loss objective (MEZL) (see  Section \ref{sec:discussion} for an explanation). 

In summary, the contributions of this paper are: 
\begin{enumerate}
    \item A Bayesian reformulation of the EER framework to enable computational efficiency with deep neural networks.
    \item An empirical evaluation of the effectiveness of the proposed method, comparing it to other state of the art deep active learning approaches on seven datasets, each with and without data shift.
\end{enumerate}

We introduce our active learning setting in Section \ref{sec:setting} before deriving our method in Section \ref{sec:method}. Next, we provide
an empirical evaluation in Section \ref{sec:experiments}. Finally, we discuss the implications, context, and related work in Section \ref{sec:discussion} and conclude with Section \ref{sec:conclusion}.

\comment{\section{Introduction}

Active learning studies how adaptivity in data collection can decrease the data requirement of machine learning systems \citet{settles2009active}, and has been applied to a variety of applications such as medical image analysis where unlabeled samples are plentiful, but labeling via annotation or experimentation is expensive. In such cases, intelligently selecting samples to reduce the number of samples necessary to train a model can significantly reduce costs. A range of active learning algorithms have been proposed in recent decades \citet{lewis1994sequential,tong2001support,roy2001toward} and current research trends have shifted towards active learning for deep learning methods \citet{sener2018active,ash2019deep,gal2017deep, sinha2019variational} which present a new set of constraints that traditional methods do not always address.

Another trend in recent literature has been on addressing data shift \citet{rabanserFL2019, schrouffMaintaining2022}: the problem of deploying a model to a distribution that is different from the training distribution. This is a critical aspect of model deployment that is common in real world applications because distributional shifts can cause significant degradation in model performance. There has also been recent interest in the intersection of active learning and data shift, characterizing methods under general data shift \cite{kirsch2021Test}, and under label shift \cite{zhaoLAY21}. Active learning has been proposed as a method to mitigate the effects of data shift by identifying salient out-of-distribution samples to be added to the training data. 

In this paper we explore a model-agnostic Bayesian derivation of EER, a method which has been shown to perform well under the setting of simple classifiers such as Naive Bayes ~\cite{roy2001toward}. We extend this work by deriving a formulation that is compatible with deep neural networks and show that it performs extremely well under contemporary settings. Intuitively, Bayesian methods capture uncertainty by using a distribution over model parameters rather than a single point estimate. Our method is general in that we only assume a probabilistic Bayesian model from which we can draw samples from the model posterior conditioned on data. Leveraging this, we are able to estimate the value of a label without retraining the neural network as would be required in the original formulation of EER. Although sampling from the exact Bayesian posterior can be difficult, there are a variety of methods to approximate this distribution such as Monte Carlo dropout \citet{gal2016dropout,gal2017deep}, ensembles \citet{beluch2018power}, and Langevin dynamics \citet{welling2011bayesian}. 

Perhaps the earliest Bayesian data collection method is Expected Information Gain \citet{lindley1956measure} which is equivalent to selecting the sample with the label that has the highest mutual information with the model parameters, also known as BALD \citet{gal2017deep}. While BALD maximizes the information encoded by model parameters about the dataset, the primary goal of classification is to identify the model parameters that minimize test error. This mismatch in the objective is what motivates the EER framework \citet{roy2001toward}: choose the sample that maximally reduces the generalization ``error'' or ``loss'' in expectation. In EER, this calculation is accomplished by retraining the model for every possible candidate point, for every possible label. The original paper performs experiments on Naive Bayes, which is computationally cheap to train. However, it is well established that deep neural networks are expensive to train and thus it would be infeasible to apply EER directly.

As a result of this, we explore a computationally efficient mathematical reformulation of EER for deep active learning. Our proposal circumvents the need for retraining by utilizing Monte Carlo dropout to approximately sample from the parameter posterior, and provides a way to use these samples to calculate the EER criteria. Similar to \citet{roy2001toward}, this derivation includes two variants: one based on zero-one loss, which we refer to as minimization of zero-one loss (MEZL), and the other on log-likelihood loss, which we refer to as minimization of log-likelihood loss (MELL). Our derivation also enables us to cast EER in an information theoretic setting and compare it with other methods based on reducing the uncertainty of the target model. In particular, we find that MELL is a close cousin of BALD in terms of reducing epistemic uncertainty, but additionally removes irrelevant information which we propose as an explanation for why MELL outperforms BALD in some of the experimental results (see Section~\ref{sec:decomposition}).

In our empirical evaluation, we expand upon existing literature by characterizing the performance of MELL, MEZL, and other state of the art active learning methods under realistic data shift scenarios by evaluating them on WILDS datasets, as well as on standard datasets CIFAR10, CIFAR100, SVHN, and MNIST with induced shift. Unlike in prior work \cite{kirsch2021Test} where the goal is to avoid labeling out of distribution samples, we are interested in the realistic scenario of reducing model error on a test set that may be drawn from a different distribution than the seed set. In addition to these experiments, we also investigate the standard active learning scenario where the test and seed sets are drawn from the same distribution. In our results, MELL consistently performs better or on par with all other information theoretic approaches in both the no data shift and data shift settings, with the largest gains above BALD coming in the data shift setting. 

The only approach that compares favorably to ours is BADGE~\cite{ash2019deep}, which outperforms MELL and MEZL on two datasets under data shift: Camelyon17 and MNIST. However, MELL and MEZL require orders of magnitude less computation cost (see Appendix~\ref{sec:computationalComplexity} and are model agnostic -- they can be applied to neural networks, random forests, linear models, etc. relying only on Bayesian sampling. 

Finally, we note from the experiments that optimizing the log likelihood (MELL) performs better than zero-one loss objective (MEZL). We offer an explanation of this phenomenon in Section \ref{sec:discussion}. 

In summary, the contributions of this paper are: 
\begin{enumerate}
    \item A Bayesian reformulation of the EER framework to enable computational efficient selection of samples that will reduce the expected loss in the context of deep neural networks.
    \item An empirical evaluation of the effectiveness of the proposed method, comparing it with other state of the art deep active learning approaches on seven datasets, both with and without data shift.
\end{enumerate}

We introduce our active learning setting in Section \ref{sec:setting} before deriving our method in Section \ref{sec:method}. Next, we provide
an empirical evaluation in Section \ref{sec:experiments}. Finally, we discuss the implications, context, and related work in Section \ref{sec:discussion} and conclude with Section \ref{sec:conclusion}.
}
\comment{
\section{Introduction}

Active learning studies how adaptivity in data collection can decrease the data requirement of machine learning systems \citet{settles2009active}, and has been applied to a variety of applications such as medical image analysis where unlabeled samples are plentiful, but labeling via annotation or experimentation is expensive. In such cases, intelligently selecting samples to reduce the number of samples necessary to train a model can significantly reduce costs. A range of active learning algorithms have been proposed in recent decades \citet{lewis1994sequential,tong2001support,roy2001toward} and current research trends have shifted towards active learning for deep learning methods \citet{sener2018active,ash2019deep,gal2017deep, sinha2019variational} which present a new set of constraints that traditional methods do not always address.

Another trend in recent literature has been on addressing data shift \steve{cite more (like three or so) papers, like surveys}: the problem of deploying a model to a distribution that is different from the training distribution. This is a critical aspect of model deployment that is common in real world applications because distributional shifts can cause significant degradation in model performance and emphasize biases in the training data \steve{emphasize biases? what do we mean by this?}. There has also been recent interest in the intersection of active learning and data shift, characterizing methods under general data shift \cite{kirsch2021Test}, and under label shift \cite{zhaoLAY21}. Active learning has been proposed as a method to mitigate the effects of data shift by identifying salient out-of-distribution samples to be added to the training data. 

In this paper we explore a model-agnostic Bayesian derivation of EER, a method which has been shown to perform well under the setting of simple classifiers such as Naive Bayes ~\cite{roy2001toward}. We extend this work by deriving a formulation that is compatible with deep neural networks and show that it performs extremely well under contemporary settings. Intuitively, Bayesian methods capture uncertainty by using a distribution over model parameters rather than a single point estimate. Our method is general in that we only assume a probabilistic Bayesian model from which we can draw samples from the model posterior conditioned on data. Leveraging this, we are able to estimate the value of a label without retraining the neural network as would be required in the original formulation of EER. Although sampling from the exact Bayesian posterior can be difficult, there are a variety of methods to approximate this distribution such as Monte Carlo dropout \citet{gal2016dropout,gal2017deep}, ensembles \citet{beluch2018power}, and Langevin dynamics \citet{welling2011bayesian}. 

Perhaps the earliest Bayesian data collection method is Expected Information Gain \citet{lindley1956measure} which is equivalent to selecting the sample with the label that has the highest mutual information with the model parameters, also known as BALD \citet{gal2017deep}. While BALD maximizes the information encoded by model parameters about the dataset, the primary goal of classification is to identify the model parameters that minimize test error. This mismatch in the objective is what motivates the EER framework \citet{roy2001toward}: choose the sample that maximally reduces the generalization ``error'' or ``loss'' in expectation. In EER, this calculation is accomplished by retraining the model for every possible candidate point, for every possible label. The original paper performs experiments on Naive Bayes, which is computationally cheap to train. However, it is well established that deep neural networks are expensive to train and thus it would be infeasible to apply EER directly.

As a result of this, we explore a computationally efficient mathematical reformulation of EER for deep active learning. Our proposal circumvents the need for retraining by utilizing Monte Carlo dropout to approximately sample from the parameter posterior, and provides a way to use these samples to calculate the EER criteria. Similar to \citet{roy2001toward}, this derivation includes two variants: one based on zero-one loss, which we refer to as minimization of zero-one loss (MEZL), and the other on log-likelihood loss, which we refer to as minimization of log-likelihood loss (MELL). Our derivation also enables us to cast EER in an information theoretic setting and compare it with other methods based on reducing the uncertainty of the target model. In particular, we find that MELL is a close cousin of BALD in terms of reducing epistemic uncertainty, but additionally removes irrelevant information which we propose as an explanation for why MELL outperforms BALD in some of the experimental results (see Section~\ref{sec:decomposition}).

In our empirical evaluation, we expand upon existing literature by characterizing the performance of MELL, MEZL, and other state of the art active learning methods under realistic data shift scenarios by evaluating them on WILDS datasets, as well as on standard datasets CIFAR10, CIFAR100, SVHN, and MNIST with induced shift. Unlike in prior work \cite{kirsch2021Test} where the goal is to avoid labeling out of distribution samples, we are interested in the realistic scenario of reducing model error on a test set that may be drawn from a different distribution than the seed set. In addition to these experiments, we also investigate the standard active learning scenario where the test and seed sets are drawn from the same distribution. In our results, MELL consistently performs better or on par with all other information theoretic approaches in both the no data shift and data shift settings. 

The only approach that compares favorably to ours is BADGE~\cite{ash2019deep}, which outperforms MELL and MEZL on two datasets under data shift: Camelyon17 and MNIST. However, as shown in Figure~\ref{fig:task-times}, BADGE is multiple orders of magnitude more expensive in terms of computational cost \steve{this is a bit too defensive/in-depth for the intro, maybe just say in simpler terms that BADGE is very computationally expensive and not model-agnostic}. In fact, BADGE is so computationally inefficient that several of the experiments did not complete after 14 days. In addition, we note that BADGE is not model free \steve{define ``model free''. If I remember correctly, one of the reviewers complained that Bayesian methods use a model}, as it requires access to the last-layer activation of the neural network. Finally, we note from the experiments that optimizing the log likelihood (MELL) performs better than zero-one loss objective (MEZL). We offer an explanation of this phenomenon in Section \ref{sec:discussion}. 

In summary, the contributions of this paper are: 
\begin{enumerate}
    \item A Bayesian reformulation of the classic expected error reduction (EER) framework to enable computational efficiency with deep neural networks.
    \item An empirical evaluation of the effectiveness of the proposed method, comparing it with other state of the art deep active learning approaches on seven datasets, both with and without data shift.
\end{enumerate}

We introduce our active learning setting in Section \ref{sec:setting} before deriving our method in Section \ref{sec:method}. Next, we provide
an empirical evaluation in Section \ref{sec:experiments}. Finally, we discuss the implications, context, and related work in Section \ref{sec:discussion} and conclude with Section \ref{sec:conclusion}.

\section{Introduction}

Active learning \citet{settles2009active} studies how adaptivity in data collection can decrease the data requirement of machine learning systems. In a variety of applications such as medical image analysis, unlabeled samples are plentiful while acquiring labels via annotation or experimentation is expensive. In such cases, intelligently selecting samples for labeling can significantly reduce costs. A range of active learning algorithms have been proposed in recent decades \citet{lewis1994sequential,tong2001support,roy2001toward}. Current research trends have shifted towards active learning for deep learning methods \citet{sener2018active,ash2019deep,gal2017deep, sinha2019variational} which present a new set of constraints.

In this paper we explore a model-agnostic Bayesian framework, and derive a computationally efficient version of EER, which has been shown to perform very well in a non-deep learning method ~\cite{roy2001toward}, and which we show performs extremely well empirically in contemporary settings. Intuitively, Bayesian methods capture uncertainty by using a distribution over model parameters, rather than a single point estimate. We use that to our advantage to estimate the value of a label, without the need for retraining. Our method is general in that we only assume a probabilistic Bayesian model from which we can draw samples from the model posterior conditioned on data. Although sampling from the exact Bayesian posterior can be difficult, there are a variety of methods to produce approximate samples such as Monte Carlo dropout \citet{gal2016dropout,gal2017deep}, ensembles \citet{beluch2018power}, and Langevin dynamics \citet{welling2011bayesian}. 

Perhaps the earliest Bayesian data collection method is Expected Information Gain \citet{lindley1956measure} which is equivalent to labeling the point whose label has the highest mutual information with the model parameters, also known as BALD \citet{gal2017deep}. While this criterion is appropriate if we want to identify the model parameters, often in machine learning the model parameters are only a means to an end, namely, to minimize the test error. This gap motivates the EER framework \citet{roy2001toward}: choose the point that maximally reduces the generalization ``error'' or loss in expectation. In \citet{roy2001toward}, this calculation is accomplished by retraining the model for every possible candidate point pseudo-labeled for every possible label. While retraining for Naive Bayes, as in \citet{roy2001toward}, is computationally cheap, retraining deep neural networks is generally computationally expensive.

We revisit EER and provide a computationally efficient mathematical reformulation for deep active learning. Our proposal circumvents the need for retraining by utilizing Monte Carlo dropout to approximately sample from the parameter posterior and deriving a way to use these samples to calculate the EER criteria. Similar to \citet{roy2001toward}, this derivation includes two variants: one based on the zero-one loss (which we call MEZL), and one on the log-likelihood loss (called MELL). Our derivation enables us to cast EER in an information theoretic setting, and to formally compare it to other methods based on reducing the  uncertainty of the target model. In particular, we find MELL is a close cousin of BALD in terms of reducing epistemic uncertainty, but additionally removes irrelevant information with respect to a specific validation set, which may explain why MELL outperforms BALD in three cases (see Section~\ref{sec:decomposition}).

Our empirical evaluation shows the efficacy of MELL and MEZL against state of the art deep active learning approaches on several benchmark datasets. In our results, MELL performs consistently better or equal to all other information theoretic approaches in both the no data shift and data shift settings. The only approach that appears to compare favorably to ours in terms of consistently performing at the top is BADGE~\cite{ash2019deep}. In two experiments involving data shift, Camelyon17 and MNIST, our approach performs slightly worse. Yet, as shown in Figure~\ref{fig:task-times}, BADGE is an order of magnitude more expensive in terms of computational time. This is to the extent that some experiments timed out after 14 days of continuous computation, preventing us from performing a full comparison for all datasets. In addition, we note that BADGE is not model free, as it requires access to the last-layer activation of the neural network. We also note in our experiments that decreasing the log likelihood (MELL) performs better than decreasing the zero-one loss (MEZL). We offer an explanation of this phenomenon in Section \ref{sec:discussion}. 

In summary, the contributions of this paper are: 
\begin{enumerate}
    \item A Bayesian reformulation of the classic expected error reduction (EER) framework to enable computational efficiency with deep neural networks.
    \item An empirical evaluation of the effectiveness of the proposed method, comparing it to other state of the art deep active learning approaches on seven datasets, each with and without data shift.
\end{enumerate}

We introduce our active learning setting in Section \ref{sec:setting} before deriving our method in Section \ref{sec:method}. Next, we provide
an empirical evaluation in Section \ref{sec:experiments}. Finally, we discuss the implications, context, and related work in Section \ref{sec:discussion} and conclude with Section \ref{sec:conclusion}.
}

\comment{
We explore a model-agnostic Bayesian framework, similar to that of \citet{gal2017deep}, and derive a computationally efficient version of Expected Error Reduction (EER) which also performs extremely well empirically. Intuitively, Bayesian methods capture uncertainty by using a distribution over model parameters, rather than a single point estimate. Our method is general in that we only assume a probabilistic Bayesian model from which we can draw samples from the model posterior conditioned on data. Although sampling from the exact Bayesian posterior can be difficult, there are a variety of methods to produce approximate samples such as Monte Carlo dropout \citet{gal2016dropout,gal2017deep}, ensembles \citet{beluch2018power}, and Langevin dynamics \citet{welling2011bayesian}. 

Perhaps the earliest Bayesian data collection method is Expected Information Gain \citet{lindley1956measure} which is equivalent to labeling the point whose label has the highest mutual information with the model parameters, also known as BALD \citet{gal2017deep}. While this criterion is appropriate if we want to identify the model parameters, often in machine learning the model parameters are only a means to an end, the test error. This gap motivates the EER framework \citet{roy2001toward}: choose the point that maximally reduces the generalization ``error'' or loss in expectation. In \citet{roy2001toward}, this calculation is accomplished by retraining the model for every possible candidate point pseudo-labeled for every possible label. While retraining for Naive Bayes, as in \citet{roy2001toward}, is computationally cheap, retraining deep neural networks is generally computationally expensive.

In this paper, we revisit EER and provide a computationally efficient mathematical reformulation for deep active learning. Our proposal circumvents the need for retraining by utilizing Monte Carlo dropout to approximately sample from the parameter posterior and deriving a way to use these samples to calculate the EER criteria. Similar to \citet{roy2001toward}, this derivation includes two variants: one based on the zero-one loss (which we call MEZL), and one on the log-likelihood loss (called MELL). Our derivation enables us to cast EER in an information theoretic setting, and to formally compare it to other methods based on reducing the  uncertainty of the target model. In particular, we find MELL is a close cousin of BALD in terms of reducing epistemic uncertainty, but additionally removes irrelevant information with respect to a specific validation set, which may explain why MELL outperforms BALD in three cases (see Section~\ref{sec:decomposition}).

We empirically evaluate the efficacy of MELL and MEZL against state of the art deep active learning approaches on several benchmark datasets. In our results, MELL performs consistently better or equal to all other information theoretic approaches in both the no data shift and data shift settings. The only approach that appears to compare favorably to ours in terms of consistently performing at the top is BADGE~\cite{ash2019deep}. Indeed, two experiments involving data shift, Camelyon17 and MNIST our approach performs slightly worse. Yet, as shown in Figure~\ref{fig:task-times}, BADGE is an order of magnitude more expensive in terms of computational time. This is to the extent that some experiments timed out after 14 days of continuous computation, preventing us from performing a full comparison for all datasets. In addition, we note that BADGE is not model free, as it requires the last-layer activation of the neural network. We note in our experiments that decreasing the log likelihood (MELL) performs better than decreasing the zero-one loss (MEZL). We offer an explanation of this phenomenon in Section \ref{sec:discussion}. 

In summary, the contributions of this paper are: 
\begin{enumerate}
    \item A Bayesian reformulation of the classic expected error reduction (EER) framework to enable computational efficiency with deep neural networks.
    \item An empirical evaluation of the effectiveness of the proposed method, comparing it to other state of the art deep active learning approaches on seven datasets, each with and without data shift.
\end{enumerate}

The rest of the paper is organized as follows. We introduce our active learning setting in Section \ref{sec:setting} before deriving our method in Section \ref{sec:method}. Next, we provide
an empirical evaluation in Section \ref{sec:experiments}. Finally, we discuss the implications, context, and related work in Section \ref{sec:discussion} and conclude with Section \ref{sec:conclusion}.
}


\comment{

Bayesian active learning is a principled framework for designing adaptive data collection methods. In particular, Bayesian methods capture uncertainty by using a distribution over model parameters, rather than a single point estimate. 

A distribution over parameters enables the definition of quantities such as the mutual information between the model parameters and an ``experiment''. (cite Lindley)

Our contributions:

\begin{itemize}
    \item A reformulation of the classic expected error reduction (EER) framework into a Bayesian setting to enable computational efficiency.
    \item An information-theoretic decomposition giving intuition behind the ranking of a few methods.
    \item An explanation of why decreasing the negative log likelihood works better than decreasing the zero-one loss in the EER framework.
    \item An empirical evaluation of the effectiveness of the proposed methods.
\end{itemize}

=================

Distribution shifts are commonly encountered in machine learning applications. When deployed in real world settings, there is no guarantee that the data distribution encountered during evaluation will be equivalent to that which is used during the training process. As a result of these shifts, metrics observed during the training process may not be representative of the model's true performance once deployed \daniel{add citations for data shift}. 

Among other approaches, active learning has been used as one method for addressing this issue by supplementing the training set with new samples drawn from the target distribution. Existing methods include techniques for diversity sampling in the target distribution subspace \cite{sener2018active}, using an auxiliary classifier to identify novel samples \cite{sinha2019variational}, and entropy estimation \cite{Holub2008EntropybasedAL}. 

In this paper, we revisit the existing method of EER and extend it for active learning for deep learning by reframing it in the context of Bayesian sampling \cite{roy2001toward}. The original method is based on computing a score for each unlabeled sample such that it minimizes the expected error over the unlabeled dataset, where expected error is approximated with either log-likelihood or accuracy. However, the method is not actionable for modern deep learning applications, as it requires the retraining of the predictive model for each potential unlabeled sample. Our method, which we refer to as expected accuracy maximization (MEZL), circumvents this by utilizing Monte Carlo dropout in approximating the conditional label distribution required in computing the expected error. 

Additionally, we characterize existing active learning methods under data shift scenario. Many surveys in this field of research remain firmly within the regime where both source and target domains are assumed to be from the same distribution. However, given the prevalence of distribution shift, it is important to consider performance under realistic conditions. 

The remainder of the paper is formatted as following: section 2 describes the setting under which the experiments are performed, while section 3 describes MEZL and its variants in detail. Section 4 will provide a theoretical motivation for the methods. Lastly, we provide experimental results on a variety of datasets, both with and without distribution shift.

\begin{itemize}
    \item Distribution shifts are common in applications: large amount of labeled data from some distribution and a small amount of labels from the evaluation distribution.
    \item On one hand, problem setup is the equivalent to standard active learning plus additional non-evaluation data.
    \item On the other hand, this setting has new challenges and opportunities.
        \subitem Partially resolves the ``learning representation'' challenge
        \subitem Bigger gains from active learning for different but overlapping distributions.
        \subitem Challenge: how to gracefully reduce to standard active learning if non-evaluation data is harmful.
\end{itemize}
}

\comment{
\steve{Please start with a paragraph on why active learning is important... Look at my attempt below - needing better english etc}

Methods for adaptive data collection, also known as active learning, have been extensively studied in the literature.  There are approaches based on defining and reducing the epistemic uncertainty~\cite{}, for estimating differences between the pool of unlabelled samples and the training data~\cite{}, and many other variants.  Among the many methods, we single out the method of Expected Error Reduction (EER), as it directly estimates the impact of labelling a sample on the accuracy of a model.  The original EER algorithm~\cite{roy2001toward} is based on computing a score for each unlabeled sample such that it minimizes the expected error over the unlabeled dataset. The expected error is approximated with either log-likelihood or accuracy. By addressing directly the impact of a label on the accuracy of the model, this method brings the promise of a strategy that should improve on any other approach.  Indeed, this method was shown empirically to be extremely successful with moderate size models~cite{} \steve{need citation}. 
However, as originally presented, EER is not actionable for modern deep learning applications, as it requires the retraining of the predictive model for each potential unlabeled sample. 
}

\section{Setting}
\label{sec:setting}

We study a classification setting where we have an input space $\mathcal{X}$ and a discrete label space $\mathcal{Y} = [C] = \{1, \dots, C\}$. We wish to find a function $f: \mathcal{X} \rightarrow \mathcal{Y}$ such that the misclassification error (zero-one loss), $\mathbb{E}_{(x,y)}[f(x) \neq y]$, is small according to some data distribution. 

In this paper we consider pool-based active learning where there is an ``unlabeled'' pool $\Pool$ drawn from some distribution $D_\text{target}$ where the labels of the samples are hidden. In each of the $K$ iterations, an active learning algorithm chooses $n_\text{query}$ samples from $\Pool$, the labels are revealed, and the samples are added to the training set. Often, active learning algorithms calculate a score for each sample and choose to query the set of  $n_\text{query}$ unlabeled samples with the highest scores. We consider two settings for the initial training set $\mathcal{D}_\text{seed}$: when it is not sampled from $D_\text{target}$, and when it is sampled from $D_\text{target}$. We refer to these scenarios as active learning with and without ``data shift'' respectively. In both cases, we evaluate on samples from $D_\text{target}$.

We take the Bayesian perspective that the model parameters $\theta$ are a random variable drawn from a known prior. As is typical, we assume that the labels of samples are conditionally independent given $\theta$. At a given iteration, let $\mathcal{D}$ be the labeled data collected thus far and $\Pr(\theta | \mathcal{D})$ be the posterior. We view the remaining unlabeled pool $\{x_i\}$ as fixed but the labels as random variables $\{Y_i\}$.



\comment{

In standard supervised learning, we have access to $n$ samples from the data distribution to infer the classification function $f$. We study pool-based active learning, where there is an unlabeled pool $\Pool$ composed of elements from $\mathcal{X}$ drawn from the data distribution but with their corresponding outputs from $\mathcal{Y}$ hidden. Then, in each of $K$ iterations, an active learning algorithm chooses $n_{query}$ unlabeled points per iteration from $\Pool$ and the labels are revealed. Often, active learning algorithms choose the set of unlabeled points to be queried as the $n_{query}$ points with the highest scores $S_i$. In this way, the differences between active learning algorithms is in how the scores $S_i$ are calculated. See Algorithm \ref{alg} for pseudo-code. The initial set of labeled data $\mathcal{D}$ is usually sampled randomly from either the data distribution or a different, related distribution. In the case that the initial $\mathcal{D}$ is sampled from a different distribution, we say this problem has ``data shift'', that is, the initial data and the evaluation distribution are different. \sout{Here, we assume that the unlabeled pool, test set, and validation set are sampled from the same distribution.}

We take a Bayesian perspective so that the model parameters $\theta$ are a random variable drawn from a known prior. As is typical, we assume that the labels are conditionally independent given $\theta$. At a given iteration, let $\mathcal{D}$ be the labeled data collected thus far and $\Pr(\theta | \mathcal{D})$ be the posterior. We view the remaining unlabeled pool $\{x_i\}$ as fixed but the labels as random variables $\{Y_i\}$. 

}
\section{Method}
\label{sec:method}

The guiding principle of our reformulation of EER is to score a candidate sample by the expected reduction in loss if we sampled the candidate's label. In this section, we provide a formalization of this idea that only requires the pairwise marginals for the labels. These marginals can be approximated in a computationally efficient way using Bayesian sampling methods such as Monte Carlo dropout \citep{gal2016dropout}.

Candidates for labeling are selected from $\Pool$. Additionally, we assume access to a small secondary set of unlabeled data for validation $\Val$, which is used for the evaluation of the expected loss reduction. Note that we do not assume that the two sets $\Pool$ and $\Val$ are drawn from the same distributions. In contrast, \citet{roy2001toward} use the same set for both validation and pool of candidates. This distinction affords us more flexibility, especially in the data shift case.

Because retraining a deep neural network for every candidate sample is computationally infeasible, our aim is to approximate the expected reduction in error. We use Bayesian sampling to approximate the effect of observing the label for every candidate sample on the error over the validation dataset and add the candidate and label that precipitates the highest reduction in error to the training set. We formalize this in Equation~\ref{eq:eer-complete} below. The $\text{Score}_i$ is the difference between two terms: the first term represents the total sum of the expected loss over $\Val$, and the second term estimates the expected loss over $\Val$ when we fix the label of the candidate sample $i$. Although we do not know the ground truth label of the candidate sample, we can take the expectation with respect to the posterior to compute an approximation for each possible label. The probabilities in the equation below are implicitly conditioned on samples from the validation set $\{x_1,...,x_{n_\text{val}}\}$ and on the candidate sample $x_i$.

\begin{align}
\label{eq:eer-complete}
\text{Score}_i = \sum_{j=1}^{n_\text{val}} \min_{a\in\mathcal{A}} \mathbb{E}_{y_j \sim Y_j | \mathcal{D}} [\ell(y_j,a)]  - 
\mathbb{E}_{y_i \sim Y_i | \mathcal{D}} \left[ \sum_{j=1}^{n_\text{val}} \min_{a \in \mathcal{A}} \mathbb{E}_{y_j \sim Y_j | \mathcal{D}, Y_i=y_i} [\ell(y_j,a)] \right]
\end{align}
Here, $a$ represents the optimal prediction (relative to a particular probability distribution), $\mathcal{A}$ is the set of possible predictions, and $\ell$ is the loss. Note that the first term in Equation~\ref{eq:eer-complete} is constant with respect to the samples of $\Pool$, and thus does not have any effect on comparing samples. Thus Equation~\ref{eq:eer-principle} is a reduced form where we estimate the score based solely on the second term; the negation of the expected loss on $\Val$ after the estimating the candidate sample's label.
\begin{align}
    \label{eq:eer-principle}
    \text{Score}_i = - \mathbb{E}_{y_i \sim Y_i | \mathcal{D}} \left[ \sum_{j=1}^{n_\text{val}} \min_{a \in \mathcal{A}} \mathbb{E}_{y_j \sim Y_j | \mathcal{D}, Y_i=y_i} [\ell(y_j,a)] \right]
\end{align}
Throughout the remainder of this work, we will continue to use $i$ and $j$ to index samples in $\Pool$ and $\Val$ respectively and omit the conditioning on $\mathcal{D}$, $\{x_1,...,x_{n_\text{val}}\}$, and $x_i$ for brevity.

We remark on the following aspects of our definition of EER ($\text{Score}_i$):
\begin{itemize}
    \item Because we focus on a Bayesian approach, we are able to sidestep the model retraining requirement in EER and work with expectations.  Appendix~\ref{sec:appendixLagrange} shows that only the quantities $\Pr(Y_i = y_i)$ and $\Pr(Y_j = y_j | Y_i = y_i)$ are needed.
    \item In Equation~\ref{eq:eer-principle}, the loss is computed with respect to the validation set $\Val$ (note that the ground truth labels for the validation set are not used). This affords additional flexibility in case the $\Pool$ samples are not drawn from the \textit{target distribution}.
\end{itemize} 

In the next subsections, we derive criteria for two different losses: the negative log-likelihood loss (MELL) and the zero-one loss (MEZL).

\subsection{MELL}
\label{sec:mell-method}
To obtain the expression for MELL, we substitute the general loss, $\ell(y_j,a)$, for the (negative) log-likelihood loss, $-\log{a_{y_i}}$, into Equation~\ref{eq:eer-principle}. For the log-likelihood loss, the set of possible predictions is the probability simplex over the $C$ classes, $\mathcal{A} = \Delta_C$, and thus $a_{y_i}$ is the vector $\Pr(Y_i | x_i, D)$. We use the non-negativity of the KL divergence to find the optimal prediction for each $a_{y_i}$ as $\Pr(Y_j = \cdot \text{ } | Y_i = y_i)$ (see the proof in Appendix~\ref{sec:appendixLagrange}), and by the definition of conditional entropy we obtain the succinct expression, $\text{Score}_i = - \sum_{j=1}^{n_\text{val}} H(Y_j | Y_i)$, where $H( \cdot | \cdot )$ is the conditional entropy. Using $H(Y_j|Y_i) = H(Y_j,Y_i) - H(Y_i)$ yields the final expression:

\begin{align}
    \text{Score}_i &= n_\text{val} H(Y_i) - \sum_{j=1}^{n_\text{val}} H(Y_j, Y_i).
\end{align}

To estimate these quantities we need to compute $\Pr(Y_i = c)$ and $\Pr(Y_i = c, Y_j = c')$. Because the parameters $\theta$ render the labels conditionally independent, we can draw $T$ samples from the posterior as $\{\theta_t\}_{t=1}^T$ and perform the following unbiased Monte Carlo approximation of these probabilities:

\begin{align}
    \Pr(Y_i = c) &\approx \frac{1}{T} \sum_{t=1}^T \Pr(Y_i = c | \theta_t) \\
    \Pr(Y_i = c, Y_j = c') &\approx \frac{1}{T} \sum_{t=1}^T \Pr(Y_i = c, Y_j = c' | \theta_t) 
    = \frac{1}{T} \sum_{t=1}^T \Pr(Y_i = c | \theta_t) \Pr(Y_j = c' | \theta_t), \label{equ:probConditional}
\end{align}

Sampling parameters from the \textit{posterior distribution} can be approximately generated in a variety of ways. In our implementation, we use \textit{Monte Carlo dropout} for its computational efficiency. Another approximate method is \textit{Cyclical SG-MCMC}, proposed in \citet{zhang2019cyclical}. In our experiments we find that \textit{Monte Carlo dropout} performs better empirically than \textit{Cyclical SG-MCMC} (see Figure~\ref{sec:learningCurves:mcmc} in Appendix~\ref{sec:mcmc-sampling}). 

\subsection{MEZL}

For the zero-one loss we substitute the general loss, $\ell(y_j,a)$, for $\mathbf{1}[y_j \neq a]$ into Equation~\ref{eq:eer-principle}, where the set of possible predictions is simply the output set $\mathcal{Y} = [C]$ and the optimal prediction for $a$ is $\text{arg}\max_{c \in [C]} \Pr(Y_j = c | Y_i = y_i)$. Then using some algebra, we obtain

\begin{align}
    \text{Score}_i = \sum_{c \in [C]} \sum_{j=1}^{n_\text{val}} \max_{c' \in [C]} \Pr(Y_j = c', Y_i = c) - n_\text{val}.
\end{align}

We can ignore the $n_\text{val}$ term since it remains constant with respect to $i$. Like MELL, the MEZL criteria requires knowing the pairwise marginals of the labels which can be estimated using Monte Carlo methods and the conditional independence of the labels given the parameters, as in Equation~\ref{equ:probConditional}.

\subsection{Information decomposition}
\label{sec:decomposition}

As before, let $Y_i$ be the label of a sample $x_i$ in $\Pool$ and $Y_j$ be the label of a sample $x_j$ in $\Val$. $\theta$ is the random variable for the posterior distribution of the parameters. Using an information-theoretic identity, the total uncertainty $H(Y_i)$ of a sample in the pool can be decomposed into the epistemic uncertainty and the aleatoric uncertainty, namely $H(Y_i) = I(Y_i; \theta) + H(Y_i | \theta).$

Because we cannot reduce the aleatoric  uncertainty $H(Y_i | \theta)$, BALD judiciously minimizes the epistemic uncertainty rather than the total uncertainty. There is a similar decomposition of the mutual information $I(Y_i; \theta)$ into the interaction information $I(Y_j; Y_i; \theta)$ and the conditional mutual information $I(Y_i; \theta | Y_j)$. Although the interaction information can in general be negative, the assumption that the labels are conditionally independent given the parameters implies that $I(Y_j; Y_i; \theta) = I(Y_j; Y_i) \geq 0$. Thus,

\begin{align}\label{decomposition}
    I(Y_i; \theta) &= I(Y_i; Y_j;  \theta) + I(Y_i; \theta | Y_j) \\
    &= I(Y_i; Y_j) + I(Y_i; \theta | Y_j).
\end{align}

Intuitively, $I(Y_i; \theta | Y_j)$ is the information that is irrelevant to the prediction of the validation sample $Y_j$. The term $I(Y_i; Y_j)$ is then the only relevant information. Note that if we take the score of a pool sample $i$ as the sum of the task-relevant information with respect to all validation samples, we recover MELL. Hence MELL can be seen as an algorithm like BALD, after removing the information that is not relevant to a validation set. 

In Appendix \ref{sec:linearmodel}, we look more closely at these terms for the case of a linear Bayesian model. We show that the irrelevant information term (second term) in decomposition \ref{decomposition} dominates the BALD sampling objective and thus samples with the largest norm are selected. In contrast, MELL focuses on the samples with larger impact on the validation set.

\comment {
\section{Method}
\label{sec:method}

The general design of our formulation of \citet{roy2001toward} is to score a candidate sample by the expected reduction in loss if we had access to the candidate sample's label, and if we predicted optimally given our knowledge thus far. In this section, we provide a formalization of this idea that only requires the pairwise marginals for the labels. These can be approximated with Monte Carlo in a computationally efficient way using Bayesian sampling methods, such as Monte Carlo dropout \citep{gal2016dropout}.

\textcolor{blue}{In our setting, the data samples for labeling are selected from a pool of unlabeled candidate samples $\Pool$. Additionally, we assume that there is a distinct set of unlabeled data for validation denoted as $\Val$. The $\Val$ set is used for the evaluation of the expected loss reduction. These two distinct data sets $\Pool$ and $\Val$ might have different distributions; in contrast, \citet{roy2001toward} used same set for the validation and for the candidate pool. This distinction allows more flexibility in the data shift setting, as the candidate pool and validation set could be from different data distributions.}

\sout{Note that while \citet{roy2001toward} use the unlabeled pool for both candidates and evaluation of expected loss, we assume we have both an unlabeled pool for the candidates and an unlabeled validation set for evaluation of the expected loss. This allows more flexibility in the data shift setting, as the candidate pool and validation set could be from different data distributions.}

\textcolor{blue}{For every candidate sample, our aim is to estimate the expected reduction in error-loss if we could observe the actual label of the sample and then retrain the model by adding the sample to the training set. As expressed in the equation below, the reduction in loss is the difference between the current loss and the expected loss with respect to the optimal prediction under the condition that we know the actual label of the candidate sample.}

For the probabilities in the expectations below, we implicitly condition on the pool of unlabeled samples and the set of unlabeled validation samples in $j = \{1,\ldots,n_{val}\}$ and consider the distribution over the labels of those sample. Additionally, we implicitly condition on any previously labeled data $\mathcal{D}$. 

\textcolor{blue}{
\begin{align*}
\sum_{j=1}^{n_{val}} \min_{a\in\mathcal{A}} \mathbb{E}_{y_j \sim Y_j | \mathcal{D}} [\ell(y_j,a)]  - 
\mathbb{E}_{y_i \sim Y_i} \left[ \sum_{j=1}^{n_{val}} \min_{a \in \mathcal{A}} \mathbb{E}_{y_j \sim Y_j | \mathcal{D}, Y_i=y_i} [\ell(y_j,a)] \right],
\end{align*}
where $a$ represents the optimal prediction given our current state of information, $\mathcal{A}$ is the set of possible predictions, and $\ell$ is the loss. 
Note that the first term is equal for all candidate samples and does not have effect on ranking these samples based on expected loss reduction. So we only need to estimate the second term to rank the samples. Thus, we can estimate the score for a candidate sample based solely on the second term; the negative expected loss after the sample's observed label. Thus, the score for the $i^{th}$ sample is:}
\begin{align}
    \label{eq:eer-principle}
    S_i = - \mathbb{E}_{y_i \sim Y_i} \left[ \sum_{j=1}^{n_{val}} \min_{a \in \mathcal{A}} \mathbb{E}_{y_j \sim Y_j | \mathcal{D}, Y_i=y_i} [\ell(y_j,a)] \right],
\end{align}
Throughout this work, we will use the index $i$ for samples from the candidate pool and the index $j$ for samples from the validation set.

\sout{The reduction in loss is the difference between the current loss (with respect to an optimal prediction given the current information) and the expected (with respect to the candidate sample's label) loss (with respect to the optimal prediction if we additionally knew the label of the candidate sample). Note that the current loss is constant for all candidate points so we can take the score to simply be the negative expected loss after the candidate's label observation. The score for the $i^{th}$ sample is:}

\sout{where $a$ is intuitively the optimal prediction given our current information, $\mathcal{A}$ is the set of possible predictions, and $\ell$ is the loss. Throughout this work, we will use the index $i$ for the candidate pool and the index $j$ for the validation set.}

\todo{please review the claim of this paragraph:}
\textcolor{blue}{The score definition and computation strategy in our work is different from \citet{roy2001toward}:
\begin{itemize}
    \item Score in \ref{eq:eer-principle} is defined with respect to the loss of the optimal prediction. This is a stronger definition than the one in \citet{roy2001toward} in which the loss is with respected to a given model.
    \item In this work, we focus on a Bayesian approach in contrast to \citet{roy2001toward} which uses point estimates. Because we are able to maintain a distribution over the quantities of interest instead of point estimates, we are able to sidestep the model retraining. In Appendix \ref{sec:appendixLagrange}, we will show that it is only required to compute the quantities $\Pr(Y_i = y_i)$ and $\Pr(Y_j = y_j | Y_i = y_i)$ to evaluate the score for the common loss functions such as cross-entropy which is done quite efficiently.
    \item In \ref{eq:eer-principle}, the loss is computed with respect to the validation set $Val$. In the data shift setting, the data sample in  $\Val$ are drown from the \textit{target domain} while the $\Pool$ samples are drown from the \textit{source domain}. This is in contrast to EER, and also to most of the other methods in active learning literature such as \textbf{BALD} and \textbf{BADGE}. 
    \todo{add more explanation}
\end{itemize} 
}

\sout{Note that we assume that we have access to the probabilities $\Pr(Y_i = y_i)$ and $\Pr(Y_j = y_j | Y_i = y_i)$. This is standard in Bayesian modeling; we rely on the model to know what would happen if we chose to label a sample.}

As in \citet{roy2001toward}, in the next subsections, we derive criteria for two different losses: the negative log-likelihood loss and the zero-one loss. For the (negative) log-likelihood loss, the possible predictions are in the probability simplex over the $C$ classes, and for the zero-one loss, the set of possible predictions is in the output set $\mathcal{Y} = [C]$.  In both cases, we are maximizing the negative expected loss which is equivalent to minimizing the expected loss. Thus, we refer to these two criteria as \emph{Minimization of Expected Log-likelihood Loss} (MELL) and \emph{Minimization of Expected Zero-one Loss} (MEZL).

\subsection{MELL}

Plugging the specific loss of MELL into the general principle:

\begin{align}
    \label{eq:mell-lagrange}
    S_i &= - \mathbb{E}_{y_i \sim Y_i} \left[  \sum_{j=1}^{n_{val}} \min_{a \in \Delta_C} \mathbb{E}_{y_j \sim Y_j|Y_i=y_i} [ - \log a_{y_j} ] \right].
\end{align}

We can use the non-negativity of the KL divergence to find the optimal prediction $a$ is $\Pr(Y_j = \cdot \text{ } | Y_i = y_i)$, see the proof in Appendix ~\ref{sec:appendixLagrange}. Then, by the definition of conditional entropy,

\begin{align}
    S_i = - \sum_{j=1}^{n_{val}} H(Y_j | Y_i),
\end{align}

where $H( \cdot | \cdot )$ is the conditional entropy. Because $I(Y_i; Y_j) = H(Y_j) - H(Y_j|Y_i)$,

\begin{align}
    S_i = \sum_{j=1}^{n_{val}} I(Y_i; Y_j)  -  \sum_{j=1}^{n_{val}} H(Y_j).
\end{align}

Noting that $\sum_{j=1}^{n_{val}} H(Y_j)$ does not depend on $i$, this term does not affect the ordering of the scores of the points in the pool. Thus, the MELL criterion is equivalent to maximizing the sum of the mutual information between the candidate point and each point in the validation set. As a result, we only require the pairwise marginal probabilities of the labels to calculate the MELL criterion. Intuitively, this is because we need to know the effect of labeling one point on the confidence of another point.

For computational purposes, we note that $H(Y_j|Y_i) = H(Y_j,Y_i) - H(Y_i)$, so:

\begin{align}
    S_i &= n_{val} H(Y_i) - \sum_{j=1}^{n_{val}} H(Y_j, Y_i).
\end{align}

Because the parameters $\theta$ render the labels conditionally independent, we can draw $T$ samples from the posterior as $\{\theta_t\}_{t=1}^T$ and perform an unbiased Monte Carlo approximation of the probabilities,

\begin{align}
    \Pr(Y_i = c) &\approx \frac{1}{T} \sum_{t=1}^T \Pr(Y_i = c | \theta_t) \\
    \Pr(Y_i = c, Y_j = c') &\approx \frac{1}{T} \sum_{t=1}^T \Pr(Y_i = c, Y_j = c' | \theta_t) \\
    = \frac{1}{T} \sum_{t=1}^T &\Pr(Y_i = c | \theta_t) \Pr(Y_j = c' | \theta_t),
\end{align}

\textcolor{blue}{Sampling parameters from the \textit{posterior distribution} can be approximately generated in a variety of ways. In our implementation, we use \textit{Monte Carlo dropout} for its computational efficiency; see \citet{gal2016dropout}. An alternative is using the \textit{Cyclical SG-MCMC} method proposed in \citet{zhang2019cyclical}. We compared two methods in \ref{sec:learningCurves:mcmc}. We did not observe advantage in using \textit{Cyclical SG-MCMC} for sampling from posterior distribution vs dropout.} 
\todo{please confirm -- did we check the cyclical method? and found no difference???}
\todo{Ehsan: yes I add result in appendix}

\subsection{MEZL}

In this section, we derive the algorithm for minimization of expected zero-one loss. Plugging $\mathcal{A}$ and $\ell$ into the principle, we get,

\begin{align}
    S_i &= - \mathbb{E}_{y_i \sim Y_i} \left[ \sum_{j=1}^{n_{val}} \min_{a \in [C]} \mathbb{E}_{y_j \sim Y_j | Y_i = y_i}[ \mathbf{1}[y_j \neq a]] \right].
\end{align}
    
The optimal prediction $a$ is simply $\text{arg}\max_{c \in [C]} \Pr(Y_j = c | Y_i = y_i)$. Then, with a step of algebra,

\begin{align}
    S_i = \sum_{c \in [C]} \sum_{j=1}^{n_{val}} \max_{c' \in [C]} \Pr(Y_j = c', Y_i = c) - n_{val}.
\end{align}

We can ignore the $n_{val}$ term at the end because it is constant with respect to $i$. Therefore, like MELL, the MEZL criteria requires knowing the pairwise marginals of the labels, which can be estimated using Monte Carlo parameter samples and the conditional independence of the labels given the parameters,

\begin{align}
    \Pr(Y_j = c, Y_i = c') &\approx \sum_{t=1}^T \Pr(Y_j = c | \theta_t) \Pr(Y_i = c' | \theta_t).
\end{align}

Unfortunately, MEZL can dramatically fail for a reason we explore in Section \ref{sec:mezl_failure}. Briefly, when calculating the score for a point $i$, a point $j$ in the validation set only contributes to the $S_i$ if observing point $i$ would change the MAP prediction on point $j$. 

\subsection{Information decomposition}
\label{sec:decomposition}

As before, let $Y_i$ be the label of a sample $x_i$ in the pool, $Y_j$ be the label of a sample $x_j$ in the validation set, and $\theta$ be a random variable for the posterior distribution of the parameters. Using an information-theoretic identity, the total uncertainty ($H(Y_i)$) of a sample in the pool can be decomposed into the epistemic uncertainty ($I(Y_i; \theta)$) and the aleatoric uncertainty ($H(Y_i|\theta)$), namely $$ H(Y_i) = I(Y_i; \theta) + H(Y_i | \theta).$$


Because we cannot reduce the aleatoric uncertainty, BALD judiciously minimizes the epistemic uncertainty rather than the total uncertainty. There is a similar decomposition of the mutual information ($I(Y_i; \theta)$) into the interaction information $I(Y_j; Y_i; \theta)$ and the conditional mutual information $I(Y_i; \theta | Y_j)$. Although interaction information $I(\cdot; \cdot; \cdot)$ can in general be negative, because of the conditional independence of the labels given the parameters in our case, $I(Y_j; Y_i; \theta) = I(Y_j; Y_i) \geq 0$.

\begin{align}\label{decomposition}
    I(Y_i; \theta) &= I(Y_i; Y_j;  \theta) + I(Y_i; \theta | Y_j) \\
    &= I(Y_i; Y_j) + I(Y_i; \theta | Y_j).
\end{align}

Intuitively, $I(Y_i; \theta | Y_j)$ is the orthogonal information, the information that is irrelevant to the prediction of the validation point $Y_j$. More precisely, the orthogonal information is the information between $Y_i$ and $\theta$ if we knew $Y_j$. On the other hand, $I(Y_i; Y_j)$ is the task-relevant information, the information after the orthogonal information has been removed.

Thus, if we take the score of a pool point $i$ as the sum of the task-relevant information with respect to all validation points, we recover MELL. Put another way, MELL can be seen as BALD after removing the information that is not relevant to a validation set.

\textcolor{blue}{
In Appendix \ref{sec:linearmodel}, we look at $I(Y_i; \theta)$ and $I(Y_i; Y_j)$ for linear Bayesian model. We show that the first term is bounded and the second term goes to infinity for large enough $\|x_i\|_\theta^2= x_i^T\Sigma_\theta x_i$.
\begin{proposition}\label{prop:bound}
Let $Y= \theta x +\epsilon$ be a linear Bayesian model where $\theta \sim \mathcal{N}(\mu_\theta, \Sigma_\theta)$ and $\epsilon\sim \mathcal{N}(0, \sigma^2)$. Then, there exists constant $C$ such that
$$E_{j\sim \Val}[I(Y_i; Y_j)] \leq C$$
for all samples $x_i$. Furthermore, $E_j[I(Y_i; \theta\vert Y_j)]= o(\log(x_i^T\Sigma_\theta x_i))$ for large enough $x_i$.
\end{proposition}
In linear case, BALD selects sample $x$ with highest norm $\|x\|_\theta =x^T\Sigma_\theta x$. Assume that $\Pool$ is drawn from a continuous unbounded distribution. The selected samples by BALD $\|x^*_i\|\rightarrow\infty$  as $|\Pool|\rightarrow \infty$. So the irreverent information term (second term) in decomposition \ref{decomposition} dominants the BALD sampling objective and thus samples with the largest norm are selected.
In contrast, the MELL objective is the relevant mutual information (first term) in decomposition \ref{decomposition}, thus it is not confused samples from the pool with large magnitude. Furthermore, we observe in linear setting that this term is approximately  the average correlation between the candidate and the validation set.
Finally, note that BADGE can not distinguished between different points in linear case, since the gradient of all points are $\theta$. 
}
}

\section{Experiments}
\label{sec:experiments}
\subsection{Setup}

We experiment with active learning on 7 image classification datasets -- CIFAR10, CIFAR100, SVHN, MNIST, Camelyon17, iWildCam, and FMoW -- each in a setting of data shift and no data shift, for a total of 14 experiments. 

Each experiment follows the standard active learning setup as described in Section \ref{sec:setting}. In the no shift setting, there is no distribution shift between seed, pool, test, and validation sets. In the shift setting, pool, test, and validation come from the same distribution, while seed is from a different distribution. Please refer to Appendix~\ref{sec:datasets} for more details about the datasets and how shifts were induced in each. Section \ref{sec:appendparameters} details the experimental parameters.

We compare MELL and MEZL to several existing methods in the literature that are representative of the state of the art in active learning. We use \textbf{Random}, consisting of uniform random sampling from the pool as a baseline. We compare to three information-theoretic methods: \textbf{BALD}~\citep{gal2017deep}, where the candidates are scored by their mutual information with the parameters $I(Y_i;\theta)$ estimated using Monte Carlo dropout; \textbf{Entropy\_MC}~\citep{lewis1994sequential}, which relies on entropy-based uncertainty sampling using probabilities from Monte Carlo dropout; and \textbf{Entropy}, which also relies on  entropy-based uncertainty sampling, this time  using probabilities from the softmax output.  The formal relation of these methods to MELL was detailed in Section~\ref{sec:decomposition}. In addition, we also compare against two empirically successful methods based on diversity sampling approaches that require access to the last layer of the task model (not model-free). The first one is  \textbf{Coreset}~\citep{sener2018active}, which relies on batch sampling a group which provides representative coverage of the unlabeled pool using the penultimate layer representation. We remark that that our experimental results for \textbf{Coreset} are generated using a greedy approximation of the method for computational efficiency. The second one is \textbf{BADGE}~\citep{ash2019deep}, which chooses samples that are diverse and high-magnitude when represented in a hallucinated gradient space with respect to model parameters in the final layer. The scoring functions for these methods can be found in Section \ref{sec:score-functions}

\subsection{Results}
\label{sec:results}

\begin{table}[ht]
\caption{ Comparison between MELL and baseline methods based on the AUC of the active learning accuracy curves. A method wins against another method if the mean AUC - standard deviation is above the other method's mean AUC + standard deviation.  Otherwise, there is tie. Note that BADGE only finished running for 9 experiments.}
\label{tab:method-comparison}
\centering
\small
\begin{tabular}{ l  c  c  c c  c  c  c } \toprule
   & BADGE & BALD & Coreset & Entropy\_MC & Entropy & MEZL & Random \\ \midrule
  MELL wins & 0 & 4 & 3 & 6 & 7 & 2 & 10\\ 
  MELL ties & 7 & 10 & 11 & 8 & 6 & 12 & 4\\ 
  MELL losses & 2 & 0 & 0 & 0 & 1 & 0 & 0 \\ 
\bottomrule
\end{tabular}
\end{table}

\begin{figure*}[h]
\centering     
\subfigure[iWildCam no shift]{\label{fig:a}\includegraphics[width=45mm]{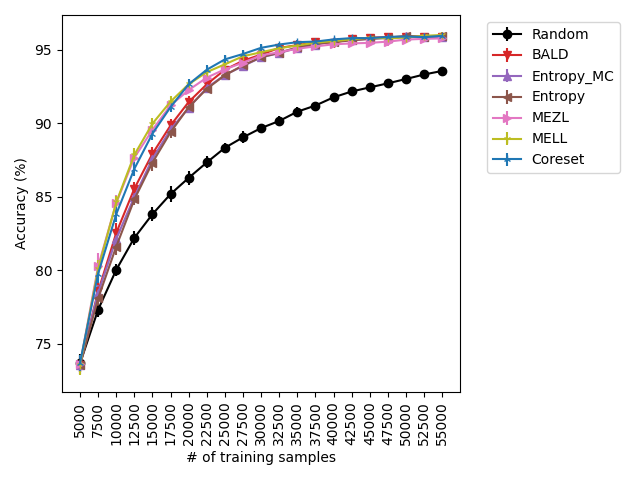}}
\subfigure[MNIST no shift]{\label{fig:b}\includegraphics[width=45mm]{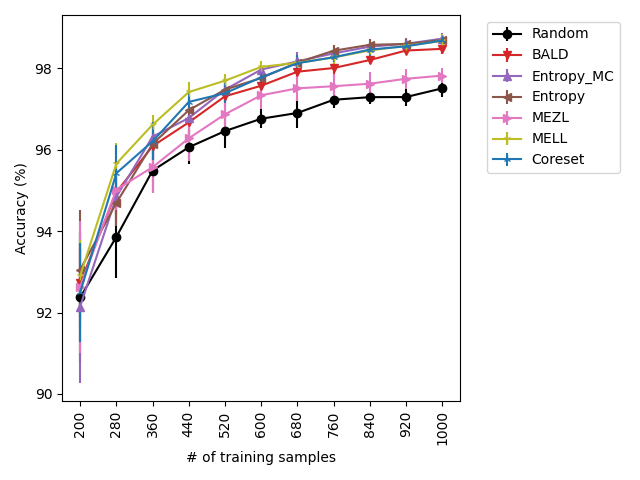}}
\subfigure[Camelyon17 no shift]{\label{fig:c}\includegraphics[width=45mm]{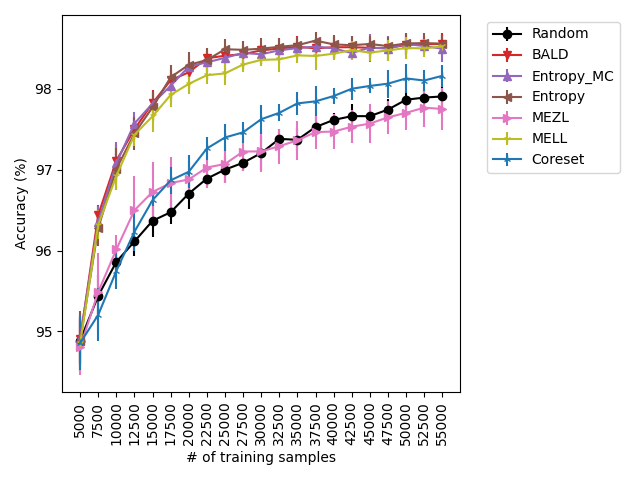}}
\subfigure[iWildCam shift]{\label{fig:d}\includegraphics[width=45mm]{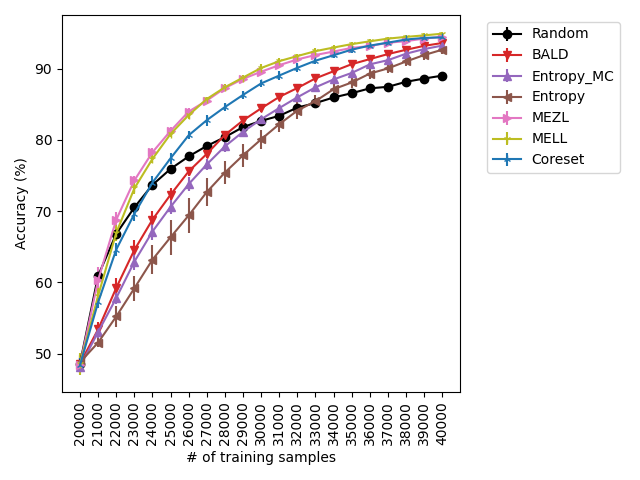}}
\subfigure[MNIST shift]{\label{fig:e}\includegraphics[width=45mm]{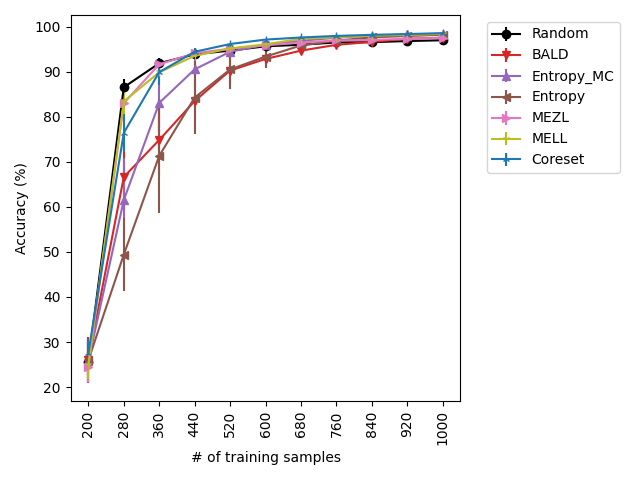}}
\subfigure[Camelyon17 shift]{\label{fig:f}\includegraphics[width=45mm]{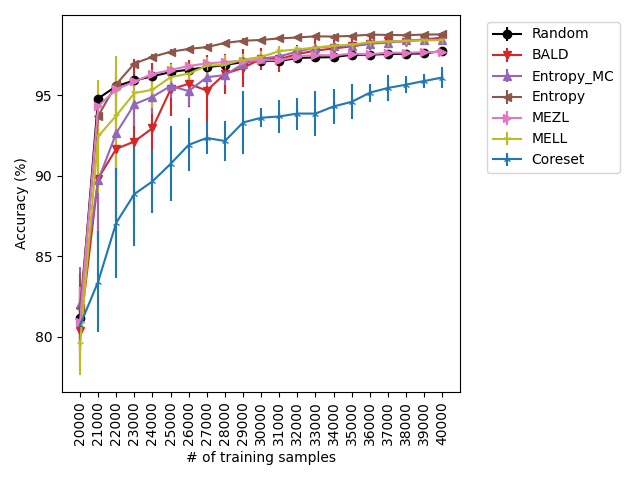}}
\caption{The accuracy curves over the test set on MNIST, Camelyon17, and iWildCam under shift and no shift settings. MELL outperforms BALD on iWildCam and MNIST in both settings. MELL outperforms Coreset on Camelyon17 in both settings and iWildCam with shift.}
\label{fig:learningCurves}
\end{figure*}

Our experiments find that MELL is the best performer or tied with best on 12 out of the 14 experiments, suggesting that MELL performs well across a variety of domains. According to Table~\ref{tab:method-comparison}, MELL never performs worse than BALD and Coreset,   outperforming BALD on 4 experiment and Coreset on 3. We show the active learning curves for the experiments where MELL outperformed BALD or Coreset in Figure \ref{fig:learningCurves}. 

For the cases where MELL outperforms BALD, we find that the performance gap is particularly wide for the data shift cases. We believe this is because BALD selects samples which maximize the information gain about the model posterior, which may not be particularly helpful for the target distribution, whereas MELL selects samples with the specific criteria of reducing error on the target distribution. 

\begin{figure}[htp]
\centering
\subfigure[Camelyon17 shift]{%
    \centering
  \includegraphics[width=45mm]{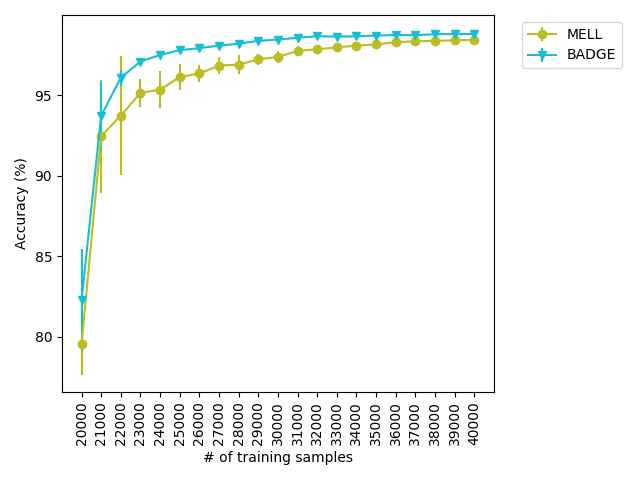}%
}
\subfigure[MNIST shift]{%
    \centering
  \includegraphics[width=45mm]{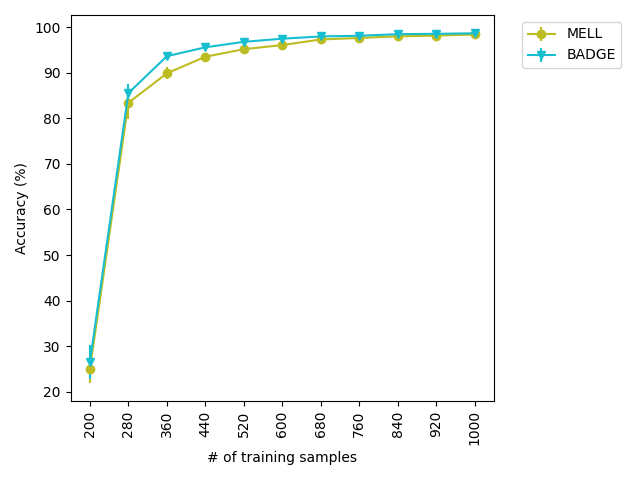}%
}
\subfigure[Camelyon17 task length]{%
\centering
  \includegraphics[width=45mm]{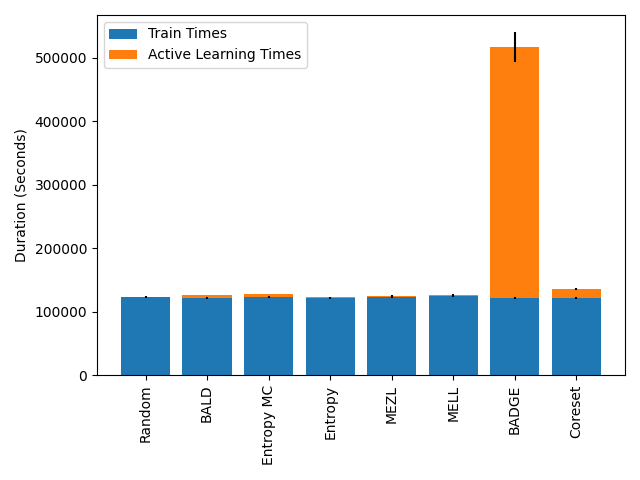}%
}
    \caption{(a) Accuracy curves and (b) total amount of time for MELL and BADGE on Camelyon17 with shift. The total time is averaged over 10 trials for each active learning strategy, and the black error bars represent the standard deviation.}
    \label{fig:task-times}
\end{figure}

In the case of data shift (see Table~\ref{tab:aucs-shift}), we have two exceptions: in MNIST, {\bf BADGE} outperforms MELL, and in Camelyon17 both {\bf BADGE} and {\bf Entropy} outperform MELL. We first note that {\bf Entropy} under-performs with respect to MELL in 7 other settings.  {\bf BADGE} on the other hand, performs on par with MELL in all the cases where the computation finishes. For iWildCam and FMoW, the computations required by {\bf BADGE} didn't finish after 14 days of execution. Indeed, {\bf BADGE} achieves its performance at a steep cost in terms of computation time.  Figure~\ref{fig:task-times} shows the exact cost for the case of Camelyon17 with shift.  Note that most of the overhead is on the active learning method itself and not on the training of the task model. This behavior is consistent in all of the experiments.   We offer some further insight in terms of computational complexity in Appendix~\ref{sec:computationalComplexity}.

In addition, we point out that it has been noted that Camelyon17 is an outlier among datasets. From \citet{miller2021accuracy}: ``One possible reason for the high variation in accuracy [on Camelyon17] is the correlation across image patches. Image patches extracted from the same slides and hospitals are correlated because patches from the same slide are from the same lymph node section, and patches from the same hospital were processed with the same staining and imaging protocol. In addition, patches in [Camelyon17] are extracted from a relatively small number of slides.'' Because of this test time dependence between images (in addition to train time dependence), \citet{miller2021accuracy} notes the presence of ``instabilities in both training and evaluation''.

\comment{Our claim on aleatoric and epistemic uncertainty is based on the fact that cases where uncertainty sampling methods perform on-par with BALD indicates that there is likely little to no aleatoric uncertainty present in those datasets, since BALD is equivalent to Entropy without the aleatoric uncertainty, as discussed in \ref{sec:decomposition}.  This is the case on iWildCam, SVHN, and Camelyon17. }

\comment{and 
As shown in Tables ~\ref{tab:saturation-no-shift-table} and Table ~\ref{tab:saturation-shift-table}, there is no alternative method that excels in all situations. The referenced tables are calculated by computing a threshold that is a percentage of the maximum achieved accuracy across all methods for a particular dataset. Then, saturation points are determined as the first budget level at which the method meets or surpasses the threshold level.}

MEZL performs on par with random selection or significantly underperforms the best methods on MNIST, CIFAR10, SVHN, and Camelyon17. 
This underperformance is due to the fact that MEZL is only sensitive to validation samples that are near the decision boundary. 
A further discussion of this phenomenon is in Section \ref{sec:mezl_failure}.

\comment{For the experiments with shift present, we begin by observing that MELL and MEZL significantly outperform all other methods on iWildCam with natural shift induced by splitting data by the location of where the images were captured Figure ~\ref{fig-3}. Notably, BALD and the uncertainty methods perform significantly worse relative to MELL and MEZL, whereas they were on-par for the iWildCam with no shift experiment. All other datasets show the same pattern -- MELL is the winner on SVHN along with uncertainty sampling and BALD, and is comparable to baseline methods for CIFAR10, CIFAR100, and FMoW. Remarkably, uncertainty sampling without Monte Carlo sampling is the clear winner on Camelyon17 with shift. 


While BADGE performs best or on par with the best active learning strategies for every dataset we ran it on, BADGE takes significantly longer to run than the other strategies. Figure \ref{fig:task-times} shows the mean time that each experiment on $\text{CIFAR10\_shift}$ took to complete, averaged over 10 trials for each strategy. For every dataset, BADGE is the slowest strategy by orders of magnitude. }

To summarize, we find that MELL outperforms or is comparable to existing state of the art methods for active learning in a variety of situations. There is no other method that performs consistently in all scenarios at the same level of computational efficiency. Our experiments hint that MELL may be appropriate for a wide range of tasks and might partially eliminate the need to make a decision about which active learning method to use. Active learning accuracy curves for all experiments and tables containing the AUCs of these experiments can be found in Appendix \ref{sec:full-results}.


\comment{Using the experimental setup discussed above, we present a selection of observed results in \ref{fig:1}; additional results are included in the appendix. In all dataests that were evaluated, we found that MELL either outperforms or is on-par with state of the art methods. On CIFAR10 (\ref{fig-1}) and SVHN, we found that all methods perform similarly, significantly outperforming random selection. MEZL outperforms random selection, but underperforms relative to the remaining methods the reasons for which are discussed in 4.3.  For CIFAR100, all dropout methods are on-par with random selection, while both uncertainty sampling methods underperform. \ref{fig-2} shows that in low data regimes for MNIST, MELL outperforms all other methods. Interestingly, uncertainty methods outperform the other dropout methods in the low data scenario as well. 
\julia{Let's move the figure legends inside the plots to save space}

With the iWilds datasets we characterized the performance of the methods with shift present. \ref{fig-3} shows that MELL and MEZL outperform all other methods. For FMoW, we see once again that MELL and MEZL outperform the baseline methods, while performing on-par with BALD. Results on Camelyon are an outlier in that uncertainty sampling outperforms all other methods. However, it is well-documented that the dataset is unusual in that the samples are not IID. \steve{citation for this?}

Our results suggest that MELL represents a significant advancement in the field of active learning, providing a way to accurately approximate the estimated error reduction. In all cases, MELL displays either improved or commensurate performance compared with state of the art methods. }

\section{Discussion}
\label{sec:discussion}

A result of the information theoretic basis of our scores, plus the use of Bayesian estimates, is that the only requirement for the computation of MELL and MEZL is the estimation of the pairwise marginal probabilities of labels. Thus, these scores can be applied to deep neural networks, random forests, linear models, etc., with very little modification. This clearly constrasts with methods such as Coreset and BADGE which make use of the last-layer activation of the neural network.  In that precise sense, our approach is \textit{model free}.

In addition, we characterized the differences between (a) entropy-based uncertainty sampling, (b) the expected information gain score (BALD), and (c) minimization of expected log-likelihood loss (MELL), in a precise information-theoretic light that explains the formal relationship between these methods. We remark that these methods are myopic in the way they score each sample independent of the others and greedily rank the outcomes. In Appendix~\ref{sec:myopic}, we discuss the possibility of extending MELL to a non-myopic algorithm.

\comment{In this work, we formalize the expected error reduction (EER) principle (see Eq.~(\ref{eq:eer-principle})) and derive two active learning scoring criteria from this general, conceptually clean principle.
\textcolor{blue}{While we assume that the model is probabilistic and predict $P(Y|x)$, our algorithms are \emph{model-free} in that they do not use any structure of the model in addition to be probabilistic base}; they could be used with any probabilistic model, whether neural networks or random forests. In fact, MELL and MEZL don't even require a parametric model; they only require the pairwise marginal probabilities of labels. In contrast, coreset \citep{sener2018active} and BADGE \citep{ash2019deep} are not model-free since they make use of the last-layer activation of the neural network. In addition, we characterized the differences between (a) entropy-based uncertainty sample, (b) expected information gain method known as BALD, and (c) minimization of expected log-likelihood loss (MELL), in a precise information-theoretic light that explains the relative ordering of these methods. \textcolor{blue}{In \ref{sec:myopic} , we discuss the possibility of extending “model free” MELL to non-myopic algorithm.}}


An important difference between our work and the original EER framework is that we make use of an unlabeled validation set rather than relying on the unlabeled pool of samples. When there is no data shift, this is a small difference. Knowledge of the labels of the validation set is not required, and we could create the validation set by subsampling the unlabeled pool before the algorithm runs. However, in settings where the unlabeled pool is from a different distribution than the test set and we don't have label access to or much data in the test set, this is an important distinction. It affords us the flexibility to focus the scoring function at a specific set, with the expectation that the algorithm will specialize to the implicit distribution in that set.


\subsection{Failure mode of MEZL}
\label{sec:mezl_failure}

MEZL performs significantly worse than MELL on some datasets because MEZL is only sensitive to validation samples near the decision boundary. More precisely, if observing the candidate label $Y_i$ does not change the prediction $Y_j$ on a validation sample, then the zero-one loss reduction is $0$.

\begin{align*}
    \text{LossReduction} = \sum_{j=1}^{n_\text{val}} \min_{a \in [C]} \mathbb{E}_{y_j \sim Y_j}[\mathbf{1}[y_j \neq a]]
    - \mathbb{E}_{y_i \sim Y_i} \left[ \sum_{j=1}^{n_\text{val}} \min_{a \in [C]} \mathbb{E}_{y_j \sim Y_j | Y_i = y_i}[ \mathbf{1}[y_j \neq a]] \right].
\end{align*}

If the prediction $a$ does not depend on the value $y_i$, then we can bring the expectation of $Y_i$ inside the $\min_{a \in [c]}$. Then, by the law of total expectation, 
the two terms are equal and there is no reduction in the zero-one loss on label $j$. Intuitively, this is a problem because observing some candidates in the pool could increase the confidence on many validation samples without necessarily changing the prediction of any of these samples. These candidates would be regarded as useless by MEZL, although they are useful \textit{in the long term} since reducing uncertainty over several iterations can reduce zero-one loss. Perhaps counter-intuitively, directly minimizing the next-step expected negative log-likelihood loss yields lower zero-one loss over many steps than directly minimizing the next-step zero-one loss.
\section{Conclusion}
\label{sec:conclusion}

In this work, we re-examine the EER algorithm \citep{roy2001toward} from a Bayesian and information theoretic perspective. Our method avoids the requirement of frequent model retraining, required in the original algorithm, which is infeasible in current deep learning models. Instead we make use of estimation techniques such as dropout that make Bayesian active learning strategies computationally efficient. Empirically, the algorithms we propose are effective on standard benchmark datasets against state of the art methods in the literature.  

We performed experiments with datasets under conditions of data shift, which have not been thoroughly examined in the active learning literature. We remark that the proper examination of active learning in conditions of shift demands further study.  In particular, there are a variety of questions related to designing and adapting active learning methods to the data shift setting that take advantage of the information afforded by the validation sets.  We leave these questions for future research, having established results in this work as initial benchmarks.  
\bibliography{bibliography.bib}
\bibliographystyle{neurips_2022}

\comment{
\section*{Checklist}
\begin{enumerate}

\item For all authors...
\begin{enumerate}
  \item Do the main claims made in the abstract and introduction accurately reflect the paper's contributions and scope?
    \answerYes{}
  \item Did you describe the limitations of your work?
    \answerYes{} In section \ref{sec:myopic}, we discuss how MELL and MEZL are motivated by the best decision for labeling a single sample rather than the optimal strategy for a batch of samples. 
  \item Did you discuss any potential negative societal impacts of your work?
    \answerNA{}
  \item Have you read the ethics review guidelines and ensured that your paper conforms to them?
    \answerYes{}
\end{enumerate}

\item If you are including theoretical results...
\begin{enumerate}
  \item Did you state the full set of assumptions of all theoretical results?
    \answerYes{} See assumptions in \ref{sec:setting} for assumptions about the setting and \ref{sec:linearmodel} for assumptions on work relating to linear Bayesian models.
        \item Did you include complete proofs of all theoretical results?
    \answerYes{} The derivations of MELL and MEZL can be found in \ref{sec:method} and the proof of \ref{prop:bound} in the Appendix.
\end{enumerate}

\item If you ran experiments...
\begin{enumerate}
  \item Did you include the code, data, and instructions needed to reproduce the main experimental results (either in the supplemental material or as a URL)?
    \answerNo{} All datasets are publicly available and experimental parameters needed to reproduce main results can be found in \ref{sec:appendparameters}. However, code is not provided.
  \item Did you specify all the training details (e.g., data splits, hyperparameters, how they were chosen)?
    \answerYes{} Training details are in \ref{sec:appendparameters}.
        \item Did you report error bars (e.g., with respect to the random seed after running experiments multiple times)?
    \answerYes{} Error bars calculated with 10 random seeds are on all figures and tables.
        \item Did you include the total amount of compute and the type of resources used (e.g., type of GPUs, internal cluster, or cloud provider)?
    \answerNo{} However, we include a plot of the task execution length in Figure \ref{fig:task-times} and a discussion of the computational complexity in Section \ref{sec:computationalComplexity}
\end{enumerate}

\item If you are using existing assets (e.g., code, data, models) or curating/releasing new assets...
\begin{enumerate}
  \item If your work uses existing assets, did you cite the creators?
    \answerYes{} In Section \ref{sec:datasets} we cite the data creators and in Section \ref{sec:appendparameters} we cite the creator of the model architecture we use.
  \item Did you mention the license of the assets?
    \answerNo{}{} The datasets we used are all public and common in research settings. All datasets are cited in the paper. They all have licenses that allow for non-commercial research.
  \item Did you include any new assets either in the supplemental material or as a URL?
    \answerNA{}
  \item Did you discuss whether and how consent was obtained from people whose data you're using/curating?
    \answerNA{}
  \item Did you discuss whether the data you are using/curating contains personally identifiable information or offensive content?
    \answerNA{}
\end{enumerate}

\item If you used crowdsourcing or conducted research with human subjects...
\begin{enumerate}
  \item Did you include the full text of instructions given to participants and screenshots, if applicable?
    \answerNA{}
  \item Did you describe any potential participant risks, with links to Institutional Review Board (IRB) approvals, if applicable?
    \answerNA{}
  \item Did you include the estimated hourly wage paid to participants and the total amount spent on participant compensation?
    \answerNA{}
\end{enumerate}

\end{enumerate}
}


\appendix

\section{Appendix}
\subsection{Log likelihood loss optimal prediction}
\label{sec:appendixLagrange}

In Section \ref{sec:mell-method}, it was noted that the optimal prediction probability $a$ is $\Pr(Y_j = \cdot | Y_i = y_i)$. Let $\vec{p}\in \Delta_C$ be the vector representation of $\Pr(Y_j = \cdot | Y_i = y_i)$; in other words, $\vec{p}_c = \Pr(Y_j = c | Y_i = y_i)$. So,

\begin{align}
    \mathbb{E}_{y_j \sim Y_j | Y_i = y_i}[-\log a_{y_j}] &= \sum_{c=1}^C \vec{p}_c (-\log a_c)
\end{align}

To show that $a=\vec{p}$ is the optimal prediction, we must show its value is always equal or less than the value of any other $a' \in \Delta_C$. Note that the KL divergence is always non-negative,

\begin{align}
    \sum_{c=1}^C \vec{p}_c  \log  \left( \frac{\vec{p}_c}{a'_c} \right) &\geq 0 \\
    \sum_{c=1}^C \vec{p}_c (-\log a'_c) &\geq \sum_{c=1}^C \vec{p}_c (-\log \vec{p}_c) \\
\end{align}

\subsection{Computational complexity}
\label{sec:computationalComplexity}
For MELL and MEZL, the amount of computation required for each candidate sample $i$ and validation sample $j$ is proportional to $T C^2$, where $T$ is the number of posterior samples and $C$ is the number of classes. Thus the leading order term in the active learning algorithm (not including the train time) is $O(K n_\text{pool} n_\text{val} T C^2)$. We can decrease the computational complexity dependence on $n_\text{pool}$ and $n_\text{val}$ by subsampling at each active learning iteration. For BADGE \citep{ash2019deep}, the main cost comes from the $k$-means++ algorithm run $K$ times (once per active learning iteration) for a total computational cost of $O(K n_\text{pool} n_\text{query} d)$, where $d$ is the embedding dimension. Because $d$ may be quite large, and we cannot subsample to reduce the dependence on $n_\text{query}$, the high computational cost of running BADGE might be expected.

\subsection{Experimental Details}
\subsubsection{Score Functions}
\label{sec:score-functions}
Below, we have the scoring functions for the baseline methods. 

\textbf{Entropy}, \textbf{Entropy\_MC}, \textbf{BALD}, and \textbf{Random} assign each data sample $i$ a score, and then query the labels of the samples with the top $n_{query}$ scores.

Given a function $\text{unif}$ that samples randomly from the uniform distribution over $[0, 1]$,  \textbf{Random} has the following scoring function: $$\text{Score}^\text{Random}_i=\text{unif}()$$

For \textbf{Entropy}: $$\text{Score}^\text{Entropy}_i=-\sum_{c\in[C]}\Pr(Y_i=c|x_i, \theta^*)\log \Pr(Y_i=c|x_i, \theta^*)$$
where $\theta^*$ is the trained model parameters.\\
For \textbf{Entropy\_MC}: $$\text{Score}^\text{Entropy\_MC}_i=-\sum_{c\in[C]}\Pr(Y_i=c|x_i, \mathcal{D})\log \Pr(Y_i=c|x_i, \mathcal{D})$$
where $$\Pr(Y_i=c|x_i, \mathcal{D}) \approx \frac{1}{T}\sum_{t=1}^T \Pr(Y_i=c|x_i, \theta_t)$$
For \textbf{BALD}: $$\text{Score}^\text{BALD}_i=\text{Score}^\text{Entropy\_MC}_i + \frac{1}{T}\sum_{t=1}^T(\sum_{c\in[C]}\Pr(Y_i=c|x_i, \theta_t)\log \Pr(Y_i=c|x_i, \theta_t))$$

\textbf{BADGE} and \textbf{Coreset} choose samples in batches. 

For \textbf{BADGE}~\cite{ash2019deep}, gradient embeddings are computed according to:
$$g_x=\frac{\partial}{\partial\theta_{\text{out}}}\ell_{\text{CE}}(f(x;\theta),\hat{y}(x))$$
where $\theta_\text{out}$ is the model parameters at the last layer, $f(x;\theta)$ is the output of model $\theta$, and $\hat{y}(x)$ is the hypothetical label on $x$. The k-means++ seeding algorithm is then run on these gradient embeddings to choose samples.

\textbf{Coreset} seeks to solve the following problem:
$$\min_{B:|B|\leq n_\text{query}}\max_{i\in \mathcal{U}}\min_{j\in B\cup \mathcal{D}}\Delta(x_i, x_j)$$
In other words, choose a batch of $n_\text{query}$ data samples, such that the maximum distance between data samples in the unlabeled set $\mathcal{U}$ and their closest center is minimized. Since computing this is NP-Hard, a k-greedy center approach is first used to initialize batch $B$, and a mixed integer program is run as a sub-routine to iteratively approach the optimal \citet{sener2018active}.

\subsubsection{Datasets}
\label{sec:datasets}
\textbf{CIFAR10} and \textbf{CIFAR100} \citep{krizhevsky2009learning} are standard image classification datasets. 
\textbf{SVHN} \citep{netzer2011reading} and \textbf{MNIST} \citep{lecun1998mnist} are digit recognition datasets, with the former containing color data from house numbers and the latter with black and white handwritten digits.
\textbf{iWildCam} \citep{beery2020iwildcam}, \textbf{Camelyon17} \citep{bandi2018detection}, and \textbf{FMoW} \citep{christie2018functional} are drawn from the WILDS repository of datasets \citep{koh2021wilds}. 
\textbf{iWildCam} contains images of wildlife captured using camera traps around the world.
\textbf{Camelyon17} contains microscope images of tissue samples taken from various patients, with the goal of identifying whether the image contains metastasized cancer cells. 
Finally, \textbf{FMoW} contains satellite imagery with the task of classifying the type of building or land represented in the image.   

For each dataset, we conduct one experiment with and one experiment without data shift. In the no shift setting, the initial training (seed), validation, unlabeled, and test sets are drawn uniformly at random from the available data (we ignore train/test/validation splits that come from the dataset). In the data shift settings, we split the dataset into a source and a target set, where source and target come from different distributions. The initial training (seed) set is sampled uniformly at random from the source set. The validation, unlabeled, and test sets are sampled uniformly at random from the target set.

We create the shift setting in CIFAR10, CIFAR100, SVHN, and MNIST by assigning the $n_\text{seed}$ images with lowest average pixel brightness to the source and all other images in the dataset to the target set. Shifted settings for Camelyon17, iWildCam, and FMoW are naturally occurring in how the datasets were collected. Using their split columns in the metadata files, we assign $\text{splits}\in\{0, 1\}$ for Camelyon17, $\text{splits}\in\{\text{val}, \text{test}\}$ for iWildCam, and $\text{splits}\in\{\text{train}\}$ for FMoW to the source set. 
We assign $\text{splits}\in\{2, 3, 4\}$ for Camelyon17, $\text{splits}\in\{\text{id\_val}, \text{id\_test}, \text{train}\}$ for iWildCam, and $\text{splits}\notin\{\text{train}\}$ for FMoW to the target set.

\subsubsection{Experimental Parameters}
\label{sec:appendparameters}

In each experiment, we initialize our training (seed), validation, unlabeled, and test sets to $n_\text{seed}$, $n_\text{val}$, $n_\text{pool}$, and $n_\text{test}$ samples respectively.
At each active learning iteration, we train a task classifier with the VGG16 architecture \citep{simonyan2014very} and an unweighted cross-entropy loss function. Then, an active learning method is used to query $n_\text{query}$ samples from the unlabeled pool to label, adding them to the training set before the next iteration. We run each experiment 10 times with different random seeds. See Table \ref{tab:training-table} for details on task model training parameters and Table \ref{tab:dataset-table} for details on experiment parameters.

\begin{table}[ht]
\caption{ A summary of hyperparameters used in our task model training. Note that the number of images processed during training (iterations multiplied by batch size) is $3000000$. In the learning rate column, the number in parenthesis is the initial learning rate for SVHN and Camelyon and the number outside is the initial learning rate for all other datasets. The learning rate is divided by 10 every 11718 iterations.}
\label{tab:training-table}
\centering
\small
\begin{tabular}{  c  c c  c  c  }\toprule
  Train Iterations & Batch Size & Learning Rate & Optimizer (Weight Decay, Momentum) & Loss\\ \midrule
  46875 & 64 & 0.01 (0.001) & SGD (5e-4, 0.9) & Cross-Entropy\\
\bottomrule
\end{tabular}
\end{table} 
\begin{table}[ht]
\caption{A summary of datasets and experimental setup parameters used in our experiments. For consistency, a two-pixel black border was added to MNIST to raise the image size from 28x28 to 32x32. $|C|$ refers to the number of classes. $L$ is the size of the subset sampled from the validation set to use in MELL and MEZL's calculations. $J$ is the size of the subset sampled from the unlabeled pool to use in MELL and MEZL's calculations. $T$ is the number of samples drawn from the posterior using Monte Carlo dropout in MELL, MEZL, Entropy MC, and BALD.}
\label{tab:dataset-table}
\centering
\begin{adjustbox}{width=1\textwidth}
\small
\begin{tabular}{ l  c  c  l  ccccccccc }\toprule
  Dataset & Image Size & $|C|$ & Shift & $n_{seed}$ & $n_{val}$ & $n_{pool}$ & $n_{query}$ & $n_{test}$ & $K$ & $L$ & $J$ & $T$ \\ \midrule
\multirow{2}{*}{CIFAR10} &\multirow{2}{*}{$32\times 32$} & \multirow{2}{*}{$10$}
& None & 5000 & 5000 & 40000 & 2500 & 10000 & 16 & 100 & 25000 & 100\\
& & & shift & 20000 & 5000 & 25000 & 1000 & 10000 & 20 & 100 & 10000 & 100\\

\midrule 
\multirow{2}{*}{CIFAR100} &\multirow{2}{*}{$32\times 32$} & \multirow{2}{*}{$100$}
& None & 5000 & 5000 & 40000 & 2500 & 10000 & 16 & 100 & 25000 & 100\\
& & & shift & 20000 & 5000 & 25000 & 1000 & 10000 & 20 & 100 & 10000 & 100\\

\midrule 
\multirow{2}{*}{SVHN} &\multirow{2}{*}{$32\times 32$} & \multirow{2}{*}{$10$}
& None & 5000 & 5000 & 50000 & 2500 & 10000 & 20 & 100 & 25000 & 100\\
& & & shift & 20000 & 5000 & 35000 & 1000 & 10000 & 20 & 100 & 10000 & 100\\

\midrule 
\multirow{2}{*}{MNIST} &\multirow{2}{*}{$32\times 32$} & \multirow{2}{*}{$10$}
& None & 200 & 5000 & 44800 & 80 & 10000 & 10 & 100 & 800 & 100\\
& & & shift &  200 & 5000 & 44800 & 80 & 10000 & 10 & 100 & 800 & 100\\


\midrule 
\multirow{2}{*}{Camelyon17} & \multirow{2}{*}{$96\times 96$} & \multirow{2}{*}{$2$}
&  None & 5000 & 5000 & 100000 & 2500 & 10000 & 20 & 100 & 25000 & 100\\ 
& & & shift & 20000 & 5000 & 100000 & 1000 & 10000 & 20  & 100 & 10000 & 100\\ 

\midrule 
\multirow{2}{*}{iWildCam} & \multirow{2}{*}{$64\times 64$} & \multirow{2}{*}{$182$} 
& None & 5000 & 5000 & 100000 & 2500 & 10000 & 20 & 100 & 25000 & 100\\ 
& & & shift & 20000 & 5000 & 100000 & 1000 & 10000 & 20 & 100 & 10000 & 100\\ 
\midrule 
\multirow{2}{*}{FMoW} & \multirow{2}{*}{$96\times 96$} & \multirow{2}{*}{$62$}
& None & 5000 & 5000 & 100000 & 2500 & 10000 & 20 & 100 & 25000 & 100\\ 
& & & shift & 20000 & 5000 & 97000 & 1000 & 10000 & 20 & 100 & 10000 & 100\\ 
\bottomrule
\end{tabular}
\end{adjustbox}
\end{table} 

\newpage
\subsection{Additional Results}
\subsubsection{Full Results}
\label{sec:full-results}

\begin{figure}
\centering     
\subfigure[MNIST no shift]{\label{fig:a}\includegraphics[width=45mm]{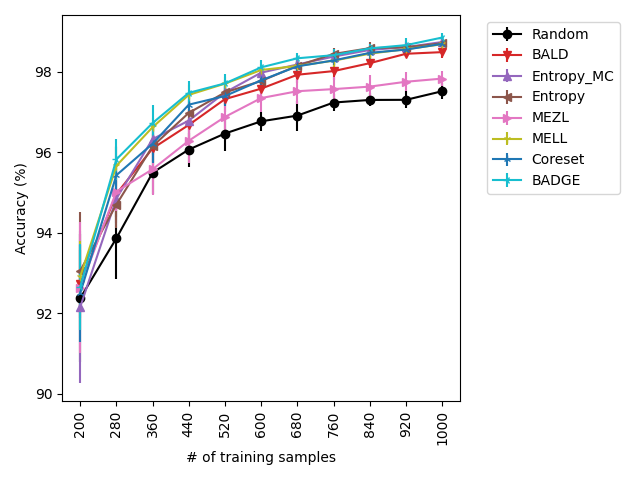}}
\subfigure[MNIST shift]{\label{fig:b}\includegraphics[width=45mm]{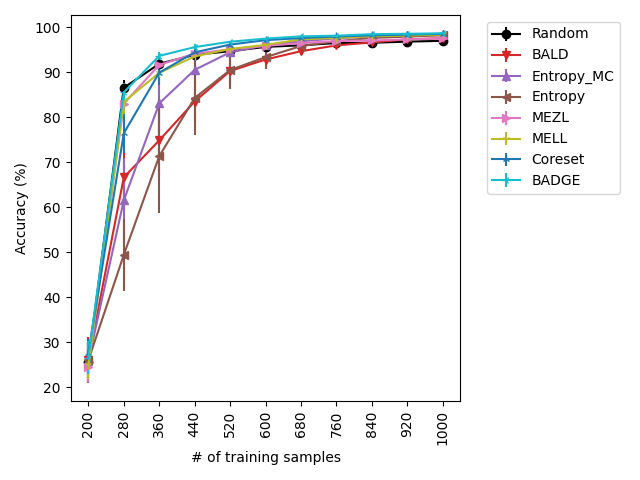}}
\caption{Active learning curves computed on the test set for all experiments.}
\label{fig:all-experiments}
\end{figure}

Below, we show the active learning accuracy curves on all datasets with and without shift for all methods in Figure \ref{fig:all-experiments}. We also show the area under the curve (AUC) of the accuracy curves in the no shift setting in Table \ref{tab:aucs-noshift} and with shift in Table \ref{tab:aucs-shift}. 

\begin{table}[ht]
\centering
\caption{\label{tab:aucs-noshift} Area under the curve (AUC) of active learning accuracy curves for each method on datasets \textbf{without shift}. Cells contain mean AUC and standard deviation across 10 trials calculated using composite Simpson's rule, as implemented in \texttt{scipy.integrate.simpson} \citep{2020SciPy-NMeth}. AUCs are normalized by the difference between the number of training samples at the last iteration and $n_\text{seed}$. "-" indicates that the experiment did not finish after 14 days.}
\begin{adjustbox}{width=1\textwidth}
\small
\begin{tabular}{ l  c c c c c c c c }
 \toprule Dataset (no shift) & MELL & MEZL & Random & BALD & Entropy MC & Entropy & Coreset & BADGE \\ \midrule
 \multirow{2}{*}{CIFAR10} &   81.74 &   81.43 & 80.65 &   81.63 &   81.65 &   81.66 &   81.58 &   81.68 \\
 &   $\pm$ 0.29 &   $\pm$ 0.40 & $\pm$ 0.36 &   $\pm$ 0.29 &   $\pm$ 0.29 &   $\pm$ 0.35 &   $\pm$ 0.37 &   $\pm$ 0.41  \\ \midrule
 
 \multirow{2}{*}{CIFAR100} &   43.96 &   43.82 &   43.83 &   43.94 & 42.83 & 42.68 &   44.30 &  -  \\
 &     $\pm$ 0.26 &      $\pm$ 0.34 &      $\pm$ 0.28 &      $\pm$ 0.34 & $\pm$ 0.32 & $\pm$ 0.34 &      $\pm$ 0.28 &   \\ \midrule 
 \multirow{2}{*}{SVHN} &      91.73 & 91.48 & 90.41 &      91.75 &      91.80 &      91.80 &      91.71 &      91.80 \\
 &      $\pm$ 0.15 & $\pm$ 0.15 & $\pm$ 0.12 &      $\pm$ 0.12 &      $\pm$ 0.15 &      $\pm$ 0.14 &      $\pm$ 0.15 &      $\pm$ 0.15  \\ \midrule 
 
  \multirow{2}{*}{MNIST} &      97.50 & 96.71 & 96.23 & 97.10 &      97.27 &      97.28 &      97.35 &      97.60 \\
 &      $\pm$ 0.11 & $\pm$ 0.27 & $\pm$ 0.28 & $\pm$ 0.18 &      $\pm$ 0.20 &      $\pm$ 0.19 &      $\pm$ 0.17 &      $\pm$ 0.18  \\ \midrule
 
 \multirow{2}{*}{iWildCam} &      92.47 &      92.24 & 88.03 & 91.84 & 91.64 & 91.57 &      92.40 &  -  \\
 &      $\pm$ 0.20 &      $\pm$ 0.15 & $\pm$ 0.24 & $\pm$ 0.21 & $\pm$ 0.16 & $\pm$ 0.18 &      $\pm$ 0.21 &   \\ \midrule 
 
 \multirow{2}{*}{Camelyon17} &      98.02 & 97.06 & 97.02 &      98.12 &      98.11 &      98.13 & 97.28 &      98.12 \\
 &      $\pm$ 0.10 & $\pm$ 0.19 & $\pm$ 0.10 &      $\pm$ 0.09 &      $\pm$ 0.11 &      $\pm$ 0.10 & $\pm$ 0.10 &      $\pm$ 0.11  \\ \midrule 
 
 \multirow{2}{*}{fMoW} &      36.43 &      36.43 & 35.10 &      36.48 & 34.90 & 34.75 &      36.67 &  -  \\
 &      $\pm$ 0.44 &      $\pm$ 0.35 & $\pm$ 0.33 &      $\pm$ 0.46 & $\pm$ 0.44 & $\pm$ 0.38 &      $\pm$ 0.33 &   \\ \bottomrule
 
\end{tabular}
\end{adjustbox}
\end{table}
\begin{table}[ht]
\centering
\caption{\label{tab:aucs-shift} Area under the curve (AUC) of active learning accuracy curves for each method on each dataset \textbf{with shift}. Cells contain mean AUC and standard deviation across 10 trials calculated using composite Simpson's rule, as implemented in \texttt{scipy.integrate.simpson} \citep{2020SciPy-NMeth}. AUCs are normalized by the difference between the number of training samples at the last iteration and $n_\text{seed}$. "-" indicates that the experiment did not finish after 14 days.}
\begin{adjustbox}{width=1\textwidth}
\small
\begin{tabular}{ l c c c c c c c c }
\toprule Dataset (shift) & MELL & MEZL & Random & BALD & Entropy MC & Entropy & Coreset & BADGE \\ \midrule
\multirow{2}{*}{CIFAR10} &    84.11 &    83.75 & 83.10 &    84.09 &    84.13 &    84.34 &    84.19 &    84.28 \\
 &    $\pm$ 0.26 &    $\pm$ 0.35 & $\pm$ 0.34 &    $\pm$ 0.23 &    $\pm$ 0.26 &    $\pm$ 0.24 &    $\pm$ 0.28 &    $\pm$ 0.26  \\ \midrule 
 
 \multirow{2}{*}{CIFAR100} &    48.51 &    48.46 &    48.16 &    48.46 &    47.93 & 47.35 &    48.19 &    48.12 \\
 &    $\pm$ 0.36 &    $\pm$ 0.34 &    $\pm$ 0.38 &    $\pm$ 0.53 &    $\pm$ 0.46 & $\pm$ 0.40 &    $\pm$ 0.35 &    $\pm$ 0.31  \\ \midrule 
 
 \multirow{2}{*}{SVHN} &    91.09 &    90.76 & 89.52 &    91.09 &    91.10 &    91.28 &    90.84 &    91.26 \\
 &    $\pm$ 0.25 &    $\pm$ 0.35 & $\pm$ 0.29 &    $\pm$ 0.28 &    $\pm$ 0.27 &    $\pm$ 0.29 &    $\pm$ 0.28 &    $\pm$ 0.25  \\ \midrule 
 
  \multirow{2}{*}{MNIST} & 91.96 & 91.71 & 91.92 & 86.03 & 88.14 & 84.06 & 91.55 &    93.32 \\
 & $\pm$ 0.52 & $\pm$ 0.42 & $\pm$ 0.32 & $\pm$ 2.28 & $\pm$ 1.77 & $\pm$ 3.54 & $\pm$ 0.53 &    $\pm$ 0.32  \\ \midrule

 \multirow{2}{*}{iWildCam} &    85.17 &    85.22 & 79.84 & 80.14 & 78.92 & 76.54 & 83.37 &  -  \\
 &    $\pm$ 0.33 &    $\pm$ 0.37 & $\pm$ 0.32 & $\pm$ 0.32 & $\pm$ 0.39 & $\pm$ 1.11 & $\pm$ 0.47 &   \\ \midrule 
 
 \multirow{2}{*}{Camelyon17} & 96.59 & 96.63 & 96.59 & 95.79 & 96.15 &    97.62 & 92.20 &    97.67 \\
 & $\pm$ 0.60 & $\pm$ 0.07 & $\pm$ 0.12 & $\pm$ 1.25 & $\pm$ 0.53 &    $\pm$ 0.10 & $\pm$ 0.54 &    $\pm$ 0.17  \\ \midrule 
 
 \multirow{2}{*}{fMoW} &    33.08 &    33.17 & 32.33 &    33.17 & 32.12 & 31.84 &    33.22 &  -  \\
 &    $\pm$ 0.27 &    $\pm$ 0.36 & $\pm$ 0.40 &    $\pm$ 0.27 & $\pm$ 0.29 & $\pm$ 0.34 &    $\pm$ 0.30 &   \\ \bottomrule
 
\end{tabular}
\end{adjustbox}
\end{table}

\begin{figure}
\centering     
\subfigure[CIFAR10 no shift]{\label{fig:a}\includegraphics[width=45mm]{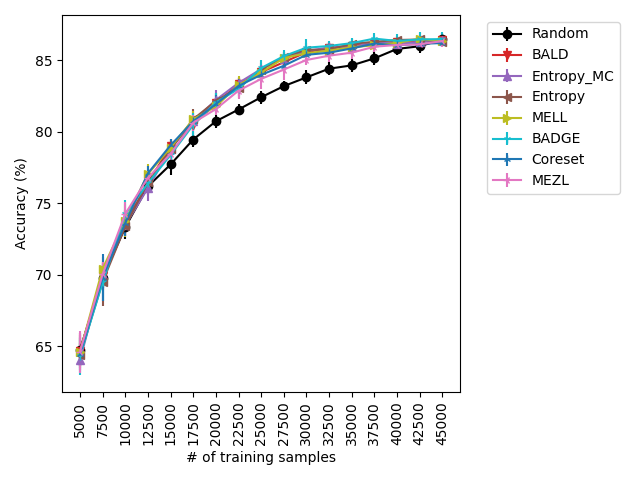}}
\subfigure[CIFAR100 no shift]{\label{fig:b}\includegraphics[width=45mm]{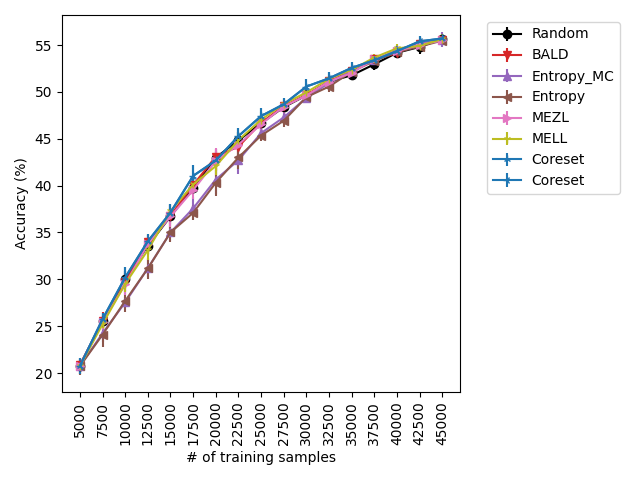}}
\subfigure[SVHN no shift ]{\label{fig:b}\includegraphics[width=45mm]{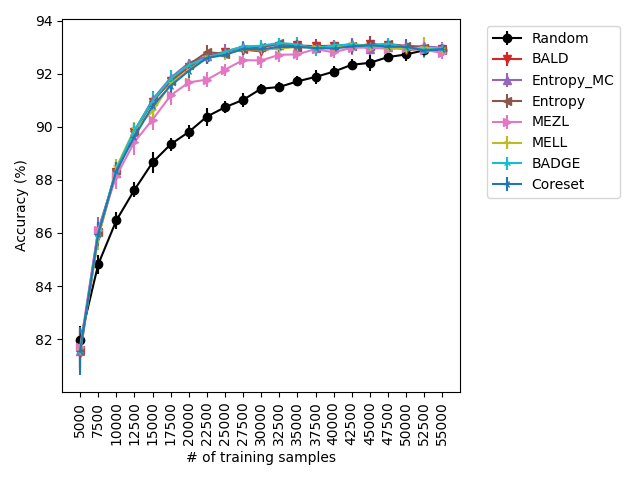}}
\end{figure}

\begin{figure}
\centering     
\subfigure[CIFAR10 shift]{\label{fig:a}\includegraphics[width=45mm]{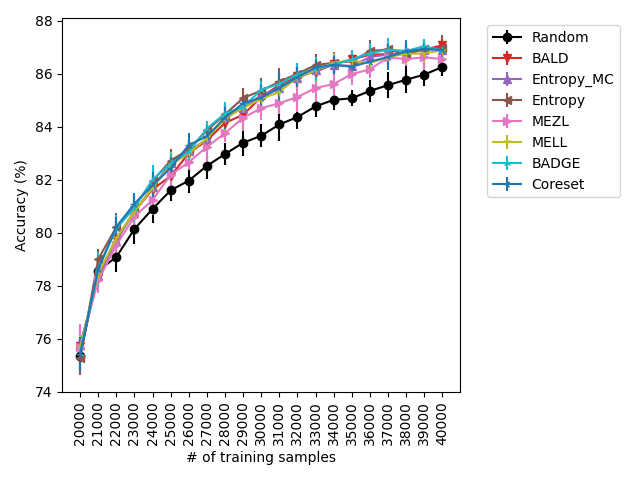}}
\subfigure[CIFAR100 shift]{\label{fig:b}\includegraphics[width=45mm]{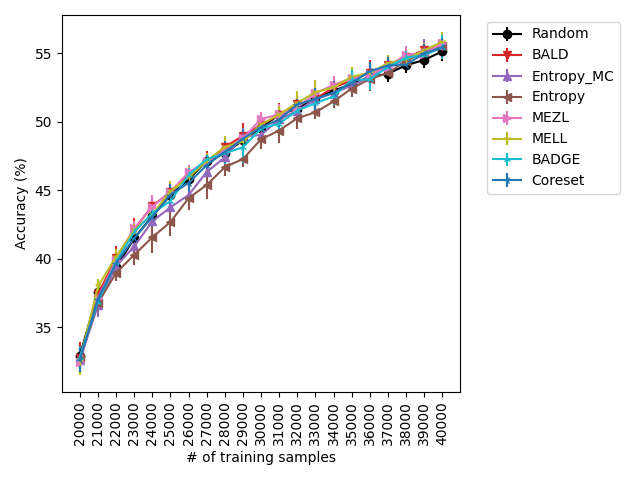}}
\subfigure[SVHN shift ]{\label{fig:b}\includegraphics[width=45mm]{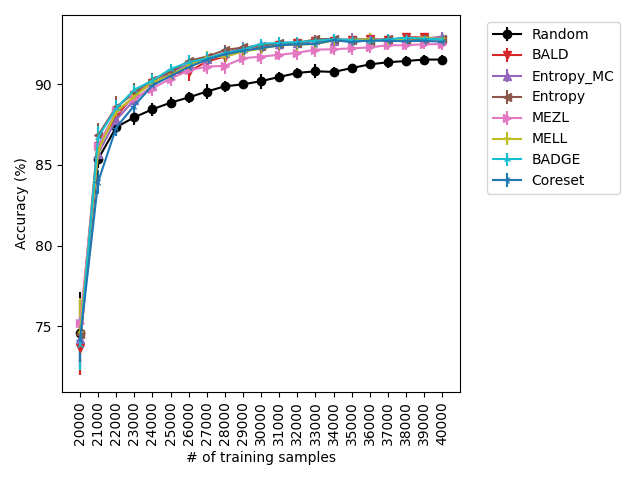}}
\end{figure}

\begin{figure}
\centering     
\subfigure[iWildCam no shift]{\label{fig:a}\includegraphics[width=45mm]{images/accuracies_on_iwildcam_noshift_accuracies.png}}
\subfigure[FMoW no shift]{\label{fig:b}\includegraphics[width=45mm]{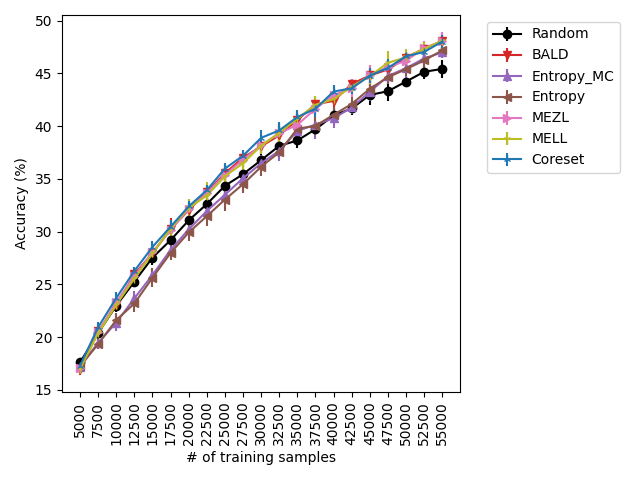}}
\subfigure[Camelyon17 no shift ]{\label{fig:b}\includegraphics[width=45mm]{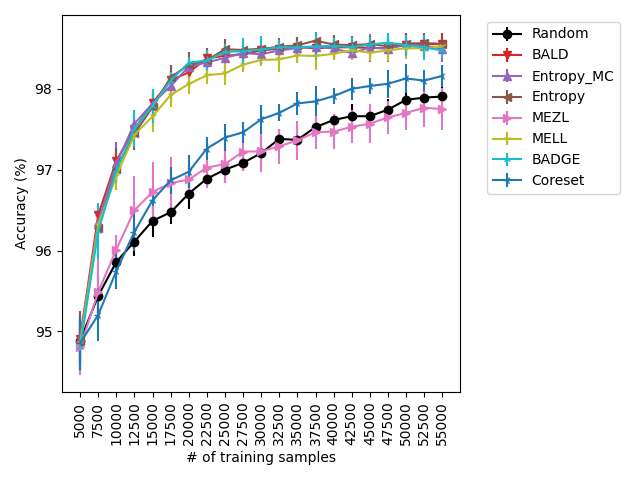}}
\end{figure}

\begin{figure}
\centering     
\subfigure[iWildCam shift]{\label{fig:a}\includegraphics[width=45mm]{images/accuracies_on_iwildcam_shift_accuracies.png}}
\subfigure[FMoW shift]{\label{fig:b}\includegraphics[width=45mm]{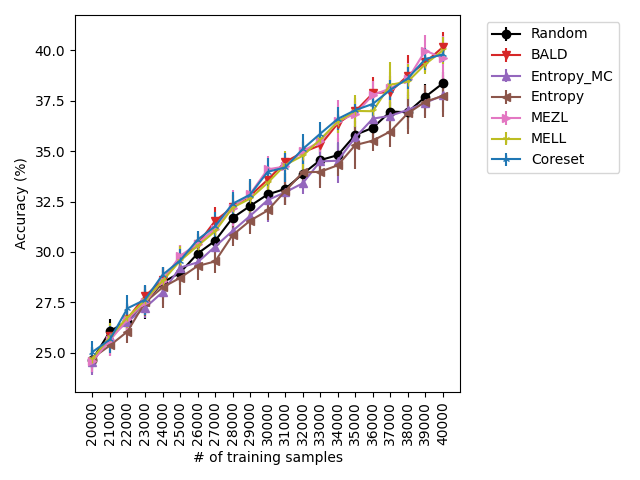}}
\subfigure[Camelyon17 shift ]{\label{fig:b}\includegraphics[width=45mm]{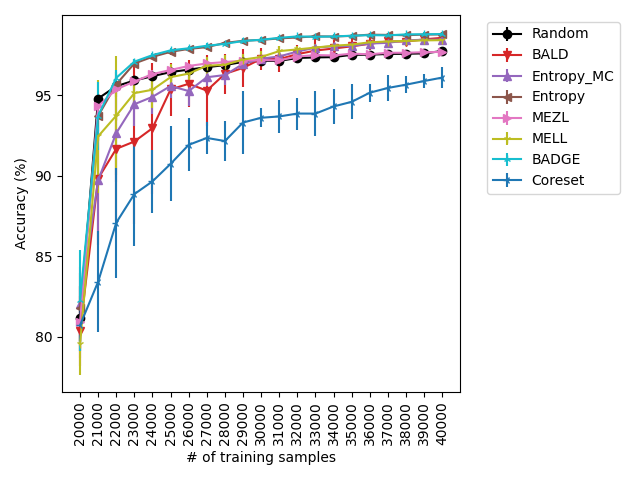}}
\end{figure}

\newpage
\subsubsection{Cyclical SG-MCMC Sampling Method}\label{sec:mcmc-sampling}

In this experiment, we compare using \textit{Cyclical SG-MCMC} as our Bayesian sampling method against \textit{Monte Carlo dropout}. We train all task models by \textit{Cyclical SG-MCMC} proposed in \citet{zhang2019cyclical}. The initial step-size is $\alpha_0=0.5$ and the number of cycles is $M=10$. We collect $10$ samples per cycle. Collected samples are used as the samples from posterior distribution in MELL\_MCMC. For all other methods, we use dropout to collect samples from the posterior distribution. According to the experiment, we did not observe any significant improvement to using \textit{Cyclical SG-MCMC} compared to \textit{drop-out} in the active learning context.

\begin{figure}[ht]
\centering
\subfigure[CIFAR10 no-shift]{\label{sec:learningCurves:mcmc:fig:a}\includegraphics[width=55mm]{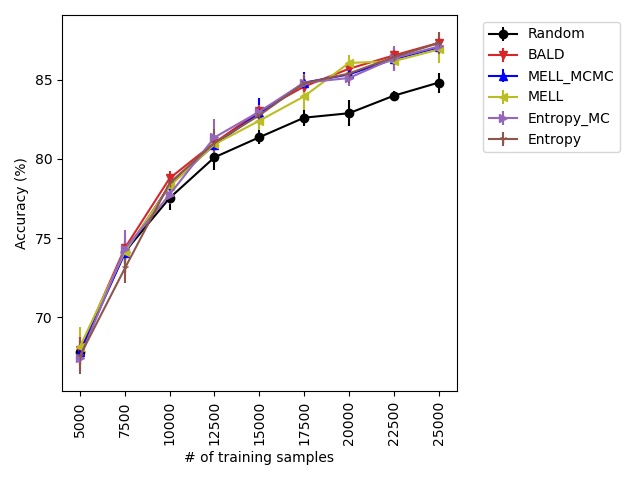}}
\subfigure[iWildCam no-shift]{\label{sec:learningCurves:mcmc:fig:b}\includegraphics[width=55mm]{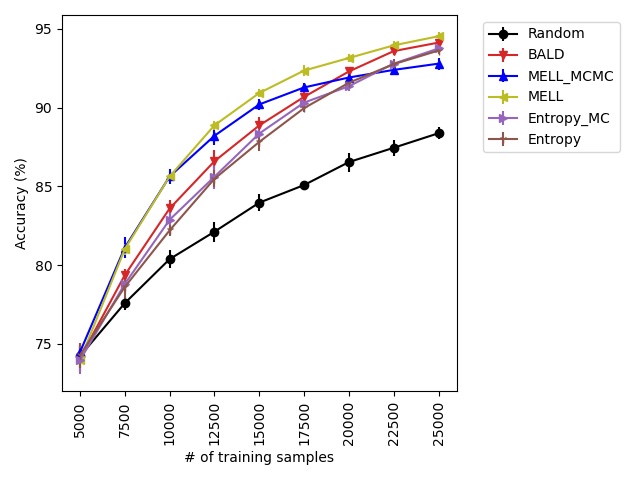}}
\caption{Cyclical SG-MCMC is used in training all methods. Posterior samples are generated by Monte Carlo dropout for all methods, except MELL\_MCMC, for which we use samples from "sampling phase" in Cyclical SG-MCMC. In both experiments, we used datasets
without data shift.}
\label{sec:learningCurves:mcmc}
\end{figure}


\subsection{General Bayesian active learning guarantees}
\label{sec:myopic}
To the best of our knowledge, finding a computationally efficient (polynomial-time) algorithm with guarantees for the model-free Bayesian active learning setting is an open challenge. All Bayesian active learning algorithms mentioned here (MELL, MEZL, BALD, Entropy) are {\em myopic}, in that they are motivated by the best decision \emph{if the budget only allowed for a single query}, and have clear failure modes. In particular consider the following simple distribution over $100$ pool samples and one validation sample. That is $\forall i \in [100]: Y_i \sim U(\{0,1\})$, and $Y_\text{val} = Y_1 \text{ XOR } Y_2.$


Suppose we wish to predict $Y_\text{val}$ by querying a subset of $\{Y_i\}_{i=1}^{100}$. In this case, the optimal strategy is to query $Y_1$ and $Y_2$ from which we can compute $Y_\text{val}$. However, existing algorithms value all possible queries equally. Because each $Y_i$ has the same distribution, uncertainty sampling with entropy will behave randomly. If we set the parameters $\theta$ for BALD to be the joint distribution, conditioning on any single variable decreases the entropy slightly, but by the same amount for each query. MEZL and MELL will also behave randomly because no query by itself gives any information about $Y_\text{val}$. 

An algorithm based on planning ahead by two queries will behave optimally for this task, but at a large computational burden. Furthermore, planning ahead by two queries will fail for a slightly more difficult distribution, such as the case where $Y_\text{val} = Y_1 \text{ XOR } Y_2 \text{ XOR } Y_3$. More generally, if an algorithm only examines $k$-wise marginals of the labeling distribution, we can always construct a distribution such that all queries appear equal to the algorithm, but the optimal strategy only requires $k+1$ queries to remove all uncertainty about the validation set.

Designing a non-myopic strategy with guarantees remains an interesting area for future research. 
We can extend expected error reduction to the batching scenario with the log likelihood loss. We wish to find a batch $B$ of size at most $k$ that maximizes:
\begin{align}\label{app:opt:prob}
\max_
{\begin{aligned}
&B\subset\Pool\\
&|B|<k
\end{aligned}} \quad f(B) = \sum_{v \in V} H( Y_v | Y_B)
\end{align}
It is straightforward to show that
$$
f(B) = \sum_{v \in \Val} \mathbb{E}_{\vec{a}}\Big[ \sum_{c \in C} \frac{\Pr(Y_B = \vec{a},Y_v = c)}{\Pr(Y_B = \vec{a})} \log \frac{\Pr(Y_B = \vec{a},Y_v = c)}{\Pr(Y_B = \vec{a})}\Big]
$$
It is possible to estimate $f(B)$ via Monte Carlo dropout. However, optimization \ref{app:opt:prob} is NP and $f(B)$ is not a sub-modular. Therefore, finding an efficient approximation algorithm for \ref{app:opt:prob} needs further investigation.

\subsection{Analysis of MELL for Bayesian Linear Model}
\label{sec:linearmodel}

Consider the simple Bayesian linear model. Here, $y= \theta^T x+\epsilon$ is a multivariate normal distribution where  $\theta \sim \mathcal{N}(\mu_\theta, \Sigma_\theta)$ and $\epsilon\sim   \mathcal{N}(0, \sigma^2)$ and $\mu_\theta\in \mathbb{R}^d$,  $\Sigma_\theta\in \mathbb{R}^{d\times d}$, and assume that $\Sigma_\theta$ is positive-definite. For a given sample $x$, the distribution of its label is $$Y_x\sim \mathcal{N}(\mu_\theta^T x, x^T\Sigma_\theta x+\sigma^2).$$ 
Therefore,
$$
H[Y_x] = \frac{1}{2}\log(2\pi e ) + \frac{1}{2}\log( x^T\Sigma_\theta x+\sigma^2).
$$
Then, we have
\begin{align}\label{app:mutual}
    I(Y_x,\theta) = H[Y_x] - E[H[Y_x|\theta]]=  \frac{1}{2}\log( x^T\Sigma_\theta x+\sigma^2) -  \frac{1}{2}\log( \sigma^2).
\end{align}
\textbf{BALD} selects the samples with highest mutual information between model parameters and the label of the sample. Therefore, BALD chooses the samples with highest $x^T\Sigma_\theta x$ for $x\in \Pool$. This  criterion is known in Bayesian Experimental Design literature; for instance \citet{chaloner1995bayesian}. Assume that the samples in pool $\Pool$ is drawn from a continuous unbounded data distribution, then $\max_{x\in \Pool} x^T\Sigma_\theta x \rightarrow \infty$ as $n_\text{pool}\rightarrow \infty$ where $n_\text{pool}=|\Pool|$. For instance, if the underlying distribution for the pool data set is Gaussian then we can use the properties of the maxima of Gaussian distribution in \citet{hartigan2014bounding} and show that\footnote{Here, $ f(x) = \Theta (g(x))$ means that there exists positive constants $c_1$, $c_2$ and $x_0$ such that $c_1 g(x) \leq f(x) \leq c_2  g(x)$ for all $x \geq x_0$.}
\begin{align*}{\max_{x\in \Pool}}x^T\Sigma_\theta x = \Theta(\log(n_\text{pool}))
\end{align*}
 which implies that if $x^*$ is the selected sample by BALD, then $\|x^*\|\rightarrow \infty$ as $n_\text{pool}\rightarrow \infty$.

\textbf{MELL} selects the samples with highest mutual information between the sample’s label and the labels of the validation set. Let $x$ be the candidate sample and $v$ is a sample drawn from validation set with corresponding labels $Y_x$ and $Y_v$. The labels for these two samples are jointly Gaussian and one can easily show that $\text{Var}(Y_x) = x^T\Sigma_\theta x +\sigma^2$,   $\text{Var}(Y_v) = v^T\Sigma_\theta v +\sigma^2$ and $\text{cov}(Y_x, Y_x)= x^T\Sigma_\theta v$. So the correlation between these two r.v. can be derived by
\begin{align}
    r(x,v) = \text{Corr}(Y_x.Y_{v}) = \frac{x^T\Sigma_\theta v }{\sqrt{(x\Sigma_\theta x +\sigma^2)(v\Sigma_\theta v +\sigma^2)}}
\end{align}

Using the mutual information formula for jointly-Gaussian distribution, we can show that
\begin{align}\label{app:mell}
E_v[I(Y_x,Y_v)] =\frac{-1}{2} E_v\Big[\log\Big(1-r(x,v)^2\Big)\Big]
\end{align}
For small enough $\frac{\Sigma_\theta}{\sigma^2}\ll 1$, 
$$E_v[I(Y_x,Y_v)]\approx
\frac{1}{2} E_v[r(x,v)^2]$$
Therefore, MELL selects the most correlated samples with the validation set, in contrast to BALD which is selecting samples with the largest norm. 

Recall from decomposition \ref{decomposition}, we have
\begin{align}\label{app:decomposition}
    I(Y_x; \theta) &=I(Y_x; Y_v) + I(Y_x; \theta | Y_v).
\end{align}

In the next proposition, we show that the first term of right hand side is bounded and the second term goes to infinity for large enough $\|x\|_\theta^2= x^T\Sigma_\theta x$ .
As we mentioned earlier, the selected samples by BALD $\|x^*_i\|\rightarrow\infty$ as $|\Pool|\rightarrow \infty$ if $\Pool$ are drawn from an unbounded distribution. So the irreverent information term (second term) in decomposition \ref{app:decomposition} dominants the BALD sampling objective and thus samples with the largest norm are selected.
In contrast, the MELL objective is the relevant mutual information (first term) in decomposition \ref{app:decomposition}, thus it is not confused by samples from the pool with large magnitude.

\begin{proposition}\label{prop:bound}
Let $Y= \theta x +\epsilon$ be a linear Bayesian model where  $\theta \sim \mathcal{N}(\mu_\theta, \Sigma_\theta)$ and $\epsilon\sim \mathcal{N}(0, \sigma^2)$ are independent. Then, there exists constant $C$ such that
$$E_{v}[I(Y_x; Y_v)] \leq C.$$
for all $x\in \Pool$. Furthermore, $E_v[I(Y_x; \theta\vert Y_v)]= \Theta(\log(x^T\Sigma_\theta x))$ for large enough $x$.
\end{proposition}
\begin{remark} We show that $$C=\frac{1}{2\sigma^2} E_{v}[v^T\Sigma_\theta v]$$
\end{remark}
\begin{proof}
Given that $Y_x = \theta^T x +\epsilon$ and $Y_v = \theta^T v +\epsilon'$, we have
\begin{align}
\text{Var}(\alpha Y_x - Y_v) &= \text{Var}((\alpha \epsilon- \epsilon') + \theta^T(\alpha x-v))\nonumber\\
&\geq \sigma^2 (1+\alpha^2)\geq \sigma^2\label{app:linearmodel:ineq}
\end{align}
for every $\alpha\in \mathbb{R}$. Here we use the independence assumption between parameters and noise. 
Letting $\alpha = \frac{\text{cov}(Y_x,Y_v)}{\text{Var}(Y_x)}$, observe that
\begin{align*}
\text{Var}(\alpha Y_x - Y_v) &= \alpha^2 \text{Var}(Y_x) -2\alpha \cdot\text{cov}(Y_x, Y_v) + \text{Var}(Y_v)\\ 
&= \frac{\text{cov}(Y_x,Y_v)^2}{\text{Var}(Y_x)} -2 \frac{\text{cov}(Y_x,Y_v)}{\text{Var}(Y_x)}\cdot \text{cov}(Y_x, Y_v) + \text{Var}(Y_v)\\
&=\text{Var}(Y_v) -  \frac{\text{cov}(Y_x,Y_v)^2}{\text{Var}(Y_x)}
\end{align*}
Using Inequality \ref{app:linearmodel:ineq}, we can conclude that  
\begin{align*}
\text{Var}(Y_v) -  \frac{\text{cov}(Y_x,Y_v)^2}{\text{Var}(Y_x)}\geq \sigma^2
\end{align*}
Observe that $\text{Var}(Y_x) = x^T\Sigma_\theta x +\sigma^2$,   $\text{Var}(Y_v) = v^T\Sigma_\theta v +\sigma^2$ and $\text{cov}(Y_x, Y_x)= x^T\Sigma_\theta v$. Thus, 
\begin{align*}
1-r(x,v)^2 = 1 -  \frac{\text{cov}(Y_x,Y_v)^2}{\text{Var}(Y_x) \text{Var}(Y_v)} \geq \frac{\sigma^2}{\text{Var}(Y_v)} 
\end{align*}
Taking $\log$ from both sides and using the fact that $\log z> 1-1/z$, we have 
\begin{align*}
\log(1-r(x,v)^2) \geq \log( \frac{\sigma^2}{\sigma^2+v^T\Sigma_\theta v}) > 1-\frac{1}{\sigma^2}(\sigma^2 + v^T\Sigma_\theta v) = 
-\frac{1}{\sigma^2}( v^T\Sigma_\theta v)
\end{align*}
By taking the expectation and using Equation \ref{app:mell}, we can conclude that
$$E_{v}[I(Y_x; Y_v)]  < C=\frac{1}{2\sigma^2} E_{v}[v^T\Sigma_\theta v]$$
for all $x\in \Pool$. The second part of the proposition is a direct result of decomposition \ref{app:decomposition} and the observation that $I(Y_x,\theta)=\Theta(\log(x^T\Sigma_\theta x))$ for linear model.
\end{proof}

\end{document}